\documentclass[runningheads]{llncs}
\usepackage{graphicx}
\usepackage{comment}
\usepackage{amsmath,amssymb} % define this before the line numbering.
\usepackage{color}
\usepackage{hyperref}
\usepackage{cleveref}
\usepackage{booktabs}
\usepackage{cuted}
\usepackage{array}
\usepackage{ragged2e}
\usepackage{colortbl}

\usepackage[export]{adjustbox} % for fixing max width of images
\usepackage{tabularx}
\usepackage{multirow}

\def\E#1{E_\textrm{#1}}
\def\L#1{L_{\textrm{#1}}}

\definecolor{notcomp}{RGB}{255, 255, 255} % white
\definecolor{comp}{RGB}{219, 244, 255} % blue

\begin{document}
% \linenumbers
\pagestyle{headings}
\mainmatter
\def\ECCVSubNumber{1303}  % Insert your submission number here
\newcommand{\anote}[1]{{\color{red}[#1]}}

\setlength{\belowcaptionskip}{-10pt}

\title{
Who Left the Dogs Out? \\
3D Animal Reconstruction with\\
Expectation Maximization in the Loop} % Replace with your title

% \begin{comment}
% \titlerunning{ECCV-20 submission ID \ECCVSubNumber} 
% \authorrunning{ECCV-20 submission ID \ECCVSubNumber} 
% \author{Anonymous ECCV submission}
% \institute{Paper ID \ECCVSubNumber}
% %\end{comment}
%******************

\def\ss#1{\vspace{-0ex}\subsubsection{#1}}

% \begin{comment}
\titlerunning{Who left the dogs out?}
% \author{Benjamin Biggs\inst{1}\orcidID{0000-1111-2222-3333} \and
% Ollie Boyne\inst{1}\orcidID{1111-2222-3333-4444} \and
% James Charles\inst{1}\orcidID{1111-2222-3333-4444} \and
% Andrew Fitzgibbon\inst{2}\orcidID{2222--3333-4444-5555}\and
% Roberto Cipolla\inst{1}\orcidID{2222--3333-4444-5555}}
\author{Benjamin Biggs\inst{1} \and
Oliver Boyne\inst{1} \and
James Charles\inst{1} \and\\
Andrew Fitzgibbon\inst{2} \and
Roberto Cipolla\inst{1}}
\authorrunning{B. Biggs et al.}
\institute{
    Department of Engineering, 
    University of Cambridge, 
    Cambridge, 
    UK
    \email{\{bjb56,ob312,jjc75,rc10001\}@cam.ac.uk} \and
    Microsoft, 
    Cambridge,
    UK
    \email{awf@microsoft.com}}
% \end{comment}
%******************
\maketitle
% https://www.overleaf.com/project/5e4fcc782813ac000121f3e4
% ~\vspace{-9mm}\\
\begin{abstract}
We introduce an automatic, end-to-end method for recovering the 3D pose and shape of dogs from monocular internet images. 
The large variation in shape between dog breeds, significant occlusion and low quality of internet images makes this a challenging problem.
We learn a richer prior over shapes than previous work, which helps regularize parameter estimation.
We demonstrate results on the Stanford Dog Dataset, an ``in-the-wild'' dataset of 20,580 dog images for which we have collected 2D joint and silhouette annotations to split for training and evaluation. 
In order to capture the large shape variety of dogs, we show that the natural variation in the 2D dataset is enough to learn a detailed 3D prior through expectation maximisation (EM).
As a by-product of training, we generate a new parameterized model (including limb scaling) SMBLD which we release alongside our new annotation dataset \emph{StanfordExtra} to the research community. Code and data are available at \url{https://sites.google.com/view/wldo}.
% I would try to get away from these -- they are from the 17th century...
% (i.e. remove them, and see if the conference people complain)
% \keywords{animal tracking, 3D morphable models, shape from silhouette}
\end{abstract}

\section{Introduction}
Animals contribute greatly to our society, in numerous ways economic and otherwise (there are more than 63 million pet dogs in the US alone~\cite{appa20}).
In consequence, there has been considerable attention in the computer vision research community to the interpretation of imagery of animals.
Although these techniques share similarities to techniques for understanding images of humans, a key difference is that obtaining labelled training data for animals is more difficult than for humans, because of the wide range of shapes and species of animals, and the difficulty of educating manual labellers in animal physiology.
\begin{figure}[t]
\setlength{\fboxsep}{0pt}%
\setlength{\fboxrule}{0pt}%

% Define left and right aligned fixed width columns
\renewcommand\tabularxcolumn[1]{m{#1}}% for vertical centering text in X column
\newcolumntype{L}[1]{>{\hsize=#1\hsize\raggedright\arraybackslash}X}%
\renewcommand\tabularxcolumn[1]{m{#1}}% for vertical centering text in X column
\newcolumntype{R}[1]{>{\hsize=#1\hsize\raggedleft\arraybackslash}X}%

\begin{tabularx}{1\textwidth}{@{} *{3}{R{0.1666}L{0.1666}}@{}p{0cm} @{}}

    \includegraphics[height=0.2\linewidth, max width=0.15\linewidth]{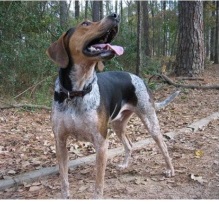} &
    \includegraphics[height=0.2\linewidth, max width=0.15\linewidth]{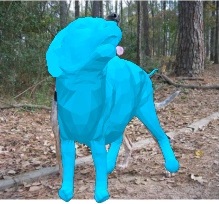} &

    \includegraphics[height=0.2\linewidth, max width=0.15\linewidth, max width=0.15\linewidth]{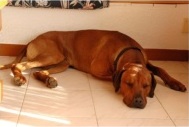} &
    \includegraphics[height=0.2\linewidth, max width=0.15\linewidth, max width=0.15\linewidth]{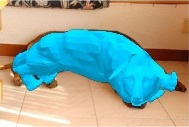} &

    \includegraphics[height=0.2\linewidth, max width=0.15\linewidth]{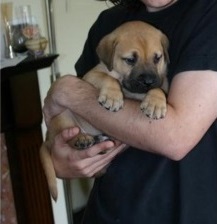} &
    \includegraphics[height=0.2\linewidth, max width=0.15\linewidth]{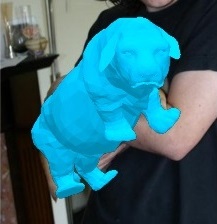} &

    \\

    \includegraphics[height=0.2\linewidth, max width=0.15\linewidth]{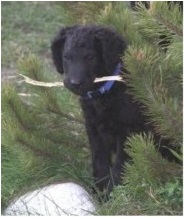} &
    \includegraphics[height=0.2\linewidth, max width=0.15\linewidth]{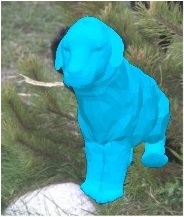} &

    \includegraphics[height=0.2\linewidth, max width=0.15\linewidth]{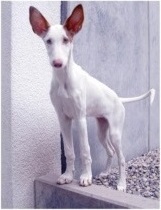} &
    \includegraphics[height=0.2\linewidth, max width=0.15\linewidth]{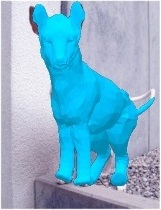} &

    \includegraphics[height=0.2\linewidth, max width=0.15\linewidth]{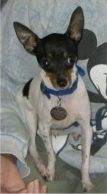} &
    \includegraphics[height=0.2\linewidth, max width=0.15\linewidth]{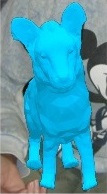} &

    \\
    \includegraphics[height=0.2\linewidth, max width=0.15\linewidth]{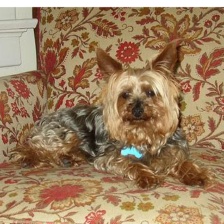} &
    \includegraphics[height=0.2\linewidth, max width=0.15\linewidth]{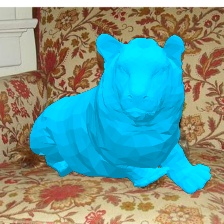} &

    \includegraphics[height=0.2\linewidth, max width=0.15\linewidth]{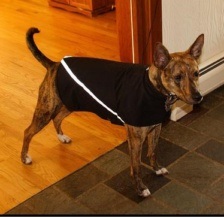} &
    \includegraphics[height=0.2\linewidth, max width=0.15\linewidth]{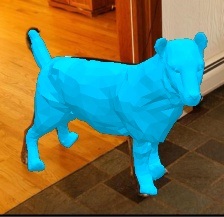} &

    \includegraphics[height=0.2\linewidth, max width=0.15\linewidth]{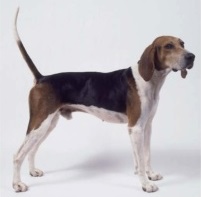} &
    \includegraphics[height=0.2\linewidth, max width=0.15\linewidth]{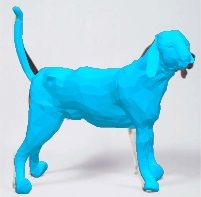}  &  

\end{tabularx}\medbreak
\caption{
\textbf{End-to-end 3D Dog Reconstruction from monocular images.}
We propose a novel method that, given a monocular image of a dog can predict a set of parameters for our SMBLD 3D dog model which is consistent with the input. We regularize learning using a multi-modal shape prior, which is tuned during training with an expectation maximization scheme.\label{fig:splash}}
\end{figure}

A particular species of interest is the dog, however it is noticeable that existing work has not yet demonstrated effective 3D reconstruction of dogs over large test sets.
We postulate that this is partially because dog breeds are remarkably dissimilar in shape and texture, presenting a challenge to the current state of the art.
The methods we propose extend the state of the art in several ways.
While each of these qualities exist in some existing works, we believe ours is the first to exhibit this combination, leading to a new state of the art in terms of scale and object diversity.
\begin{enumerate}
    \item We reconstruct pose and shape on a test set of 1703 low-quality internet images of a complex 3D object class (dogs).
    \item We directly regress to object pose and shape from a single image without a model fitting stage.
    \item We use easily obtained 2D annotations in training, and none at test time.
    \item We incorporate fitting of a new multi-modal prior into the training phase (via EM update steps), rather than fitting it to 3D data as in previous work.
    \item We introduce new degrees of freedom to the SMAL model, 
    allowing explicit scaling of subparts.
\end{enumerate}

\vspace{-1em}
\subsection{Related work}

The closest work in terms of scale is the category-specific mesh reconstruction of Kanazawa et al.~\cite{kanazawa2018birds}, where 2850 images of birds were reconstructed.  However doing so for the complex pose and shape variations of dogs required the advances described in this paper.

Table~\ref{tab:literature} summarizes previous work on animal reconstruction.
It is interesting to note that while several papers demonstrate reconstruction across species, which {\em prima facie} is a richer class than just dogs, the test-time requirements (e.g. manually-clicked keypoints/silhouette segmentations, input image quality etc.) are considerably higher for those systems.
Thus we claim that the achievement of reconstructing a full range of dog breeds, 
with variable fur length, varying shape and pose of ears, and with considerable occlusion, is a significant contribution.

\newcolumntype{L}[1]{>{\RaggedRight\hspace{0pt}}p{#1}}
\newcolumntype{R}[1]{>{\RaggedLeft\hspace{0pt}}p{#1}}

\newcommand{\awfhang}[1]{
\begin{minipage}[t]{\textwidth}% Top-hanging minipage, will align on
                               % bottom of first line
\begin{tabbing} % tabbing so that minipage shrinks to fit
\\[-\baselineskip] % Make first line zero-height
#1 % Include user's text
\end{tabbing}
\end{minipage}} % can't allow } onto next line, as {WIDEBOX}~x will not tie.

\begin{table}[t!]
{\sffamily
\scriptsize
\def\hd#1{\awfhang{#1}}
\begin{tabular}{@{}L{20mm}%Paper
|L{12mm}%Class
L{15mm}%Train
|L{15mm}%Template
L{17mm}%Video
L{17mm}%Test
|L{9mm}%Model
L{5mm}%Size
@{}}
\hd{Paper}%
&\hd{Animal\\Class}%
&\hd{Training\\requirements}%
&\hd{Template\\Model}%
&\hd{Video\\required}%
&\hd{Test Time\\Annotation}%
&\hd{Model\\Fitting}%
&\hd{Test\\Size}%
\\\hline
%%%%%%%%%%%%%%%%%%%%%%
This paper
& Dogs  % 2D Joints, Silhouettes, 3D Template, 3D Priors
& J2, S2, T3, P3
& SMAL
& No & None & No & 1703
\\\hline
%%%%%%%%%%%%%%%%%%%%%%
3D-Safari~\cite{Zuffi19Safari}        
& Zebras, horses
% 3D models (albeit synthetic), 2D Joints,  Silhouettes,  3D Priors
& M3 (albeit synthetic), J2, S2, P3
& SMAL
& 3-7 frames / animal & None & Yes & 200
\\\hline
%%%%%%%%%%%%%%%%%%%%%%
Lions and Tigers and Bears (LTB)~\cite{Zuffi:CVPR:2018} 
& MLQ
& Not trained
& SMAL
& 3-7 frames / animal & J2, S2 & Yes & 14
\\\hline
%%%%%%%%%%%%%%%%%%%%%%
3D Menagerie (3D-M)~\cite{DBLP:journals/corr/ZuffiKJB16}                
& MLQ 
& Not trained
& SMAL
& No & J2, S2 & Yes & 48 
\\\hline
%%%%%%%%%%%%%%%%%%%%%%
Creatures Great and SMAL (CGAS)~\cite{biggs2018creatures}
& MLQ
& Not trained
& SMAL
& Yes & S2 (for best results shown) & Yes & 9             \\\hline 
%%%%%%%%%%%%%%%%%%%%%%
Category Specific Mesh Reconstructions~\cite{kanazawa2018birds}
& Birds
& J2, S2
& Bird convex hull
& No & None & No & 2850          
\\\hline
%%%%%%%%%%%%%%%%%%%%%%
What Shape are Dolphins~\cite{cashman2013shape}
& Dolphins, Pigeons 
& Not trained
& Dolphin Template
& 25 frames / category & J2, S2 & Yes & 25
\\\hline
%%%%%%%%%%%%%%%%%%%%%%
Animated 3D Creatures~\cite{reinert2016animatedsketching}
& MLQ
& Not trained
& Generalized Cylinders
& Yes & J2, S2 & Yes & 15
\\\hline
\end{tabular}
}
\caption{Literature summary: Our paper extends large-scale ``in-the-wild'' reconstruction to the difficult class of diverse breeds of dogs. 
MLQ: Medium-to-large quadrupeds. J2: 2D Joints. S2: 2D Silhouettes. T3: 3D Template. P3: 3D Priors. M3: 3D Model.}
\label{tab:literature}
% \vspace{-8mm}
\end{table}

% \textbf{Monocular 3D reconstruction of human bodies}
\subsubsection{Monocular 3D reconstruction of human bodies}
The majority of recent work in 3D pose and shape recovery from monocular images tackles the special case of 3D \emph{human} reconstruction. As a result, the research community has collected a multitude of open source human datasets which provide strong supervisory signals for training deep neural networks. These include accurate 3D deformable template models~\cite{loper2015smpl} generated from real human scans, 3D motion capture datasets~\cite{ionescu2013human3,vonMarcard2018} and large 2D datasets~\cite{mscoco,Johnson10,andriluka14cvpr} which provide keypoint and silhouette annotations. 

The abundance of available human data has supported the development of successful monocular 3D reconstruction pipelines~\cite{kolotouros19convolutional,hmrKanazawa17}. Such approaches rely on accurate 3D data to build detailed priors over the distribution of human shapes and poses, and use large 2D keypoints datasets to promote generalization to ``in-the-wild'' scenarios. Silhouette data has also been shown to assist in accurate reconstruction of clothes, hair and other appearance detail~\cite{pifuSHNMKL19,alldieck2019learning}.
While the dominant paradigm in human reconstruction is now end-to-end deep learning methods, SPIN~\cite{kolotouros2019learning} show impressive improvement by incorporating an energy minimization process within their training loop to further minimize a 2D reprojection loss subject to fixed pose \& shape priors. Inspired by this innovation, we learn an iteratively-improving shape prior by applying expectation maximization during the training process.

\textbf{Monocular 3D reconstruction of animal categories.}
While animals are often featured in computer vision literature, there are still relatively few works that focus on accurate 3D animal reconstruction. 

A primary reason for this is absence of large scale 3D datasets\footnote{Released after the submission of this paper, RGBD-Dog dataset~\cite{Kearney_2020_CVPR} is the first open-source 3D motion capture dataset for dogs.} stemming from the practical challenges associated with 3D motion capture, as well as a lack of 2D data which captures a wide variety of animals. The recent Animal Pose dataset~\cite{animalpose} is one such 2D alternative, but contains significantly fewer labelled images than our new StanfordDogs dataset (4,000 compared to 20,580 in ). 
On the other hand, animal silhouette data is plentiful~\cite{mscoco,everingham2010pascal,DAVIS2017-2nd}.

%In part due to the practical challenges associated with motion capture of animal categories, there is a severe lack of 3D animal data which could be used to build accurate 3D deformable models or supervise 3D predictions~\footnote{Released after the submission of this paper, RGBD-Dog dataset~\cite{Kearney_2020_CVPR} is the first open-source 3D motion capture dataset for dogs.}. 

Zuffi et al.~\cite{DBLP:journals/corr/ZuffiKJB16} made a significant contribution to 3D animal reconstruction research by releasing SMAL, a deformable 3D quadruped model (analagous to SMPL~\cite{loper2015smpl} for human reconstruction) from $41$ scans of artist-designed toy figurines. The authors also released shape and pose priors generated from artist data. In this work we develop \emph{SMBLD}, an extension of SMAL that better represents the diverse dog category by adding scale parameters and refining the shape prior using our large image dataset.

While there have been various ``model-free'' approaches which do not rely on an initial template model to generate the 3D animal reconstruction, these techniques often do not produce a mesh~\cite{Agudo_2018_CVPR,novotny19c3dpo} or rely heavily on input 2D keypoints or video at test-time~\cite{vicente_3dv,Probst2018_ECCVa}. An exception is the end-to-end network of Kanazawa et al.~\cite{kanazawa2018birds}, although we argue that the bird category exhibits more limited articulation than our dog category.

%Early work fitted simple primitives such as cylinders to user-provided limb annotations. More recently, the
We instead focus on model-based approaches. The SMAL authors~\cite{DBLP:journals/corr/ZuffiKJB16} demonstrate fitting their deformable 3D model to quadruped species using user-provided keypoint and silhouette dataset. Lions and Tigers and Bears (LTB)~\cite{Zuffi:CVPR:2018} then demonstrated fitting to broader animal categories by incorporating multi-view constraints from video sequences. Biggs et al.~\cite{biggs2018creatures} overcame the need for hand-clicked keypoints by training a joint predictor on synthetic data. 3D-Safari~\cite{Zuffi19Safari} further improve by training a deep network on synthetic data (generated using the LTB method~\cite{Zuffi:CVPR:2018}) to recover detailed zebra shapes in the wild.

A drawback of these approaches is their reliance on a test-time energy-based optimization procedure, which is susceptible to failure with poor quality keypoint/silhouette predictions and increases the computational burden. By contrast our method requires no additional energy-based refinement, and is trained purely from single in-the-wild images. The experimental section of this paper contains a robust comparison between our end-to-end method and relevant optimization-based approaches. 

% Following recent trends in human reconstruction, Zuffi et al.\cite{Zuffi19Safari} train a deep network to recover detailed zebra shapes in the wild. Their approach relies on an energy-minization step after the initial network inference and makes use of multi-view fitting (via ~\cite{Zuffi:CVPR:2018}) to generate synthetic 3D training data. By contrast our method requires no additional energy-based refinement, and is trained purely from single in-the-wild images. We further argue that our selected dog category (120 breeds) contains significantly more diverse shape and poses than are handled here.

A major impediment to research in 3D animal reconstruction has been the lack of a strong evaluation benchmark, with most of the above methods showing only qualitative evaluations or providing quantitative results on fewer than 50 examples. To remedy this, we introduce \emph{StanfordExtra}, a new large-scale dataset which we hope will drive further progress in the field. 

\begin{figure*}[h]
    \centering
    \includegraphics[width=\textwidth]{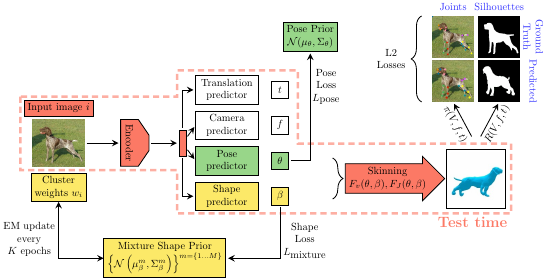}
    \caption{Our method consists of (1) a deep CNN encoder which condenses the input image into a feature vector (2) a set of prediction heads which generate SMBLD parameters for shape $\beta$, pose $\theta$, camera focal length $f$ and translation $t$ (3) skinning functions $F_v$ and $F_J$ which construct the mesh from a set of parameters, and (4) loss functions which minimise the error between projected and ground truth joints and silhouettes. Finally, we incorporate a mixture shape prior (5) which regularises the predicted 3D shape and is iteratively updated during training using expectation maximisation. At test time, our system (1) condenses the input image, (2) generates the SMBLD parameters and (3) constructs the mesh.}
    \label{fig:sys_overview_train_sup}
\end{figure*}

\section{Parametric animal model}

\def\R#1{{\mathbb{R}^{#1}}}
\def\RR#1#2{{\mathbb{R}^{#1 \times #2}}}
\def\posn{\phi}
\def\pose{\theta}
\def\npose{P}
\def\shape{\beta}
\def\scale{\kappa}
\def\trans{t}
\def\betacov{{\Sigma_{\beta}}}
\def\posecov{{\Sigma_{\pose}}}
\def\posemean{{\mu_{\pose}}}
\def\betamean{{\mu_{\beta}}}
\def\nimages{N}
\def\nshape{B}
\def\verts{\nu}
\def\nverts{V}
\def\jointselect{\mathtt{K}}
\def\njoints{J}
\def\f{f}

%\subsection{SMAL}
At the heart of our method is a parametric representation of a 3D animal mesh, which is based on the Skinned Multi-Animal Linear (SMAL) model proposed by~\cite{DBLP:journals/corr/ZuffiKJB16}. SMAL is a deformable 3D animal mesh parameterized by shape and pose. The \emph{shape}~$\shape \in \R\nshape$ parameters are PCA coefficients of an undeformed template mesh with limbs in default position. The \emph{pose}~$\pose \in \R\npose$ parameters meanwhile govern the joint angle rotations ($35 \times 3$ Rodrigues parameters) which effect the articulated limb movement. The model consists of a linear blend skinning function $F_{v}: (\pose, \shape) \mapsto V$, which generates vertex positions $V \in \RR{3889}{3}$, and a joint function $F_{J}: (\pose, \shape) \mapsto J$, which generates joint positions $J \in \RR{35}{3}$.

\subsection{Introducing scale parameters}
While SMAL has been shown to be adequate for representing a variety of quadruped types, we find that the modes of dog variation are poorly captured by the current model. This is unsurprising, since SMAL used only four dogs in its construction.

\begin{figure*}[t!]
    \centering
    \includegraphics[width=.95\linewidth]{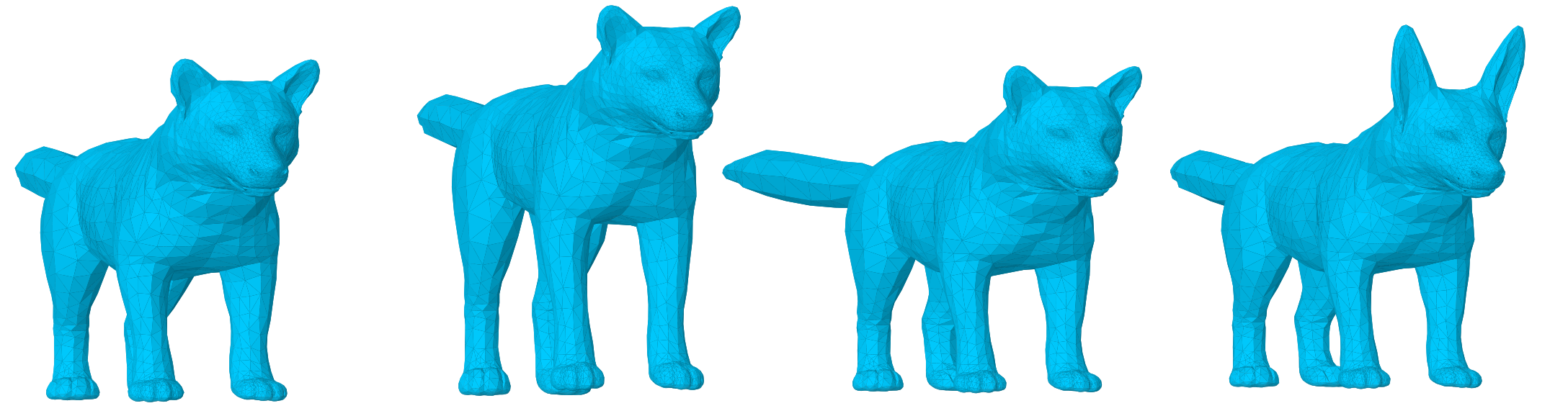}
    \caption{\textbf{Effect of varying SMBLD scale parameters}. 
    \emph{From left to right}: 
    Mean SMBLD model, 
    25\% leg elongation,
    50\% tail elongation,
    50\% ear elongation.}
    \label{fig:shape_variation}
\end{figure*}

We therefore introduce a simple but effective way to improve the model's representational power over this particularly diverse  animal category. We augment the set of shape parameters $\beta$ with an additional set $\scale$ which independently scale parts of the mesh. For each model joint, we define parameters ${\scale_x,\scale_y,\scale_z}$ which apply a local scaling of the mesh along the local coordinate $x, y, z$ axes, before pose is applied. Allowing each joint to scale entirely independently can however lead to unrealistic deformations, so we share scale parameters between multiple joints, e.g. leg lengths. The new Skinned Multi-Breed Linear Model for Dogs (SMBLD) is therefore adapted from SMAL by adding $6$ scale parameters to the existing set of shape parameters. Figure~\ref{fig:shape_variation} shows how introducing scale parameters increases the flexibility of the SMAL model. We also extend the provided SMAL shape prior (which later initializes our EM procedure) to cover the new scale parameters by fitting SMBLD to a set of $13$ artist-designed 3D dog meshes. Further details left to the supplementary.

% \subsection{Learning a unimodal 3D prior via fitting}

% Another method for improving the generalizability of the SMAL model is to improve the 3D shape prior. Such priors are typically used to ensure shape deformation remain within a realistic and anatomically plausible range. Due to the limited diversity of scans used to build the SMAL model, while the shape prior does enforce realism among deformations, it does not allow for a wide enough range to cover the set of dogs in our dataset.

% We improve the quality of the prior (and learn a prior over our new scale parameters) by fitting to a set of $13$ artist-designed 3D dog meshes, which are more varied than the original set. We apply an energy minimization scheme which aligns the SMAL vertices to each scan, under smoothing regularizers. Further details left to the supplementary.

\section{End-to-end dog reconstruction from monocular images} 

We now consider the task of reconstructing a 3D dog mesh from a monocular image. We achieve this by training an end-to-end convolutional network that predicts a set of SMBLD model and perspective camera parameters. In particular, we train our network to predict pose $\pose$ and shape $\shape$ SMBLD parameters together with translation $\trans$ and focal length $f$ for a perspective camera. A complete overview of the proposed system is shown in Figure~\ref{fig:sys_overview_train_sup}.

\vspace{-1em}
\subsection{Model architecture}

%extended with convolutional layer and an fully-connected layer 
Our network architecture is inspired by the model of 3D-Safari~\cite{Zuffi19Safari}. Given an input image cropped to (224, 224), we apply a Resnet-50~\cite{he2016deep} backbone network to encode a 1024-dimensional feature map. These features are passed through various linear prediction heads to produce the required parameters. The pose, translation and camera prediction modules follow the design of 3D-Safari, but we describe the differences in our shape module.

\vspace{-1em}
\ss{Pose, translation and camera prediction.}
These modules are independent multi-layer perceptrons which map the above features to the various parameter types. As with 3D-Safari we use two linear layers to map to a set of $35 \times 3$ 3D pose parameters (three parameters for each joint in the SMBLD kinematic tree) given in Rodrigues form. We use independent heads to predict camera frame translation $\trans_{x,y}$ and depth $\trans_{z}$ independently. We also predict the focal length of the perspective camera similarly to 3D-Safari.

\ss{Shape and scale prediction.}

Unlike 3D-Safari, we design our network to predict the set of shape parameters (including scale) rather than vertex offsets. We observe improvement by handling the standard 20 blend-shape parameters and our new scale parameters in separate linear prediction heads. We retrieve the scale parameters by $\scale = \exp{x}$ where $x$ are the network predictions, as we find predicting log scale helps stabilise early training.

\vspace{-1em}
\subsection{Training losses}

A common approach for training such an end-to-end system would be to supervise the prediction of $(\pose, \shape, \trans, \f)$ with 3D ground truth annotations~\cite{kolotouros2019learning,kanazawa18end-to-end,pavlakos18learning}. However, building a suitable 3D annotation dataset would require an experienced graphics artist to design an accurate ground truth mesh for each of 20,520 StanfordExtra dog images, a prohibitive expense.

We instead develop a method that instead relies on \emph{weak 2D supervision} to guide network training. In particular, we rely on only 2D keypoints and silhouette segmentations, are significantly cheaper to obtain.

The rest of this section describes the set of losses used to supervise the network at train time.

\ss{Joint reprojection.}
The most important loss to promote accurate limb positioning is the joint reprojection loss $\L{joints}$ which compares the projected model joints $\pi(F_{J}(\pose, \shape), \trans, \f)$ to the ground truth annotations $\hat{X}$. Given the parameters predicted by the network, we apply the SMBLD model to transform the pose and shape parameters into a set of 3D joint positions $J \in \RR{35}{3}$, and project them to the image plane using translation and camera parameters. The joint loss $L_{joints}$ is given by the $\ell_2$ error between the ground truth and projected joints:

\begin{equation}
\L{joints}(\pose, \shape, \trans, \f; \hat{X}) = \lVert \hat{X} - \pi(F_{J}(\pose, \shape), \trans, \f) \rVert_{2}
\end{equation}

Note that many of our training images exhibit significant occlusion, so $\hat{X}$ contains many invisible joints. We handle this by masking $\L{joints}$ to prevent invisible joints contributing to the loss.

\ss{Silhouette loss.}
The silhouette loss $\L{sil}$ is used to promote shape alignment between the SMBLD dog mesh and the input dog. In order to compute the silhouette loss, we define a rendering function $R: (\verts, \trans, \f) \mapsto S$ which projects the SMBLD mesh to produce a binary segmentation mask. In order to allow derivatives to be propagated through $R$, we implement $R$ using the differentiable Neural Mesh Renderer~\cite{kato2018renderer}. The loss is computed as the $\ell_2$ difference between a projected silhouette and the ground truth mask $\hat{S}$:

\begin{equation}
\L{sil}(\pose, \shape, \trans, \f; \hat{S}) = \lVert \hat{S} - R\bigl(F_{V}(\pose, \shape), \trans, \f \bigr) \rVert_{2}
\end{equation}

\ss{Priors.}
In the absence of 3D ground truth training data, we rely on priors obtained from artist graphics models to encourage realism in the network predictions. We model both pose and shape using a multivariate Gaussian prior, consisting of means $\mu_{\pose},\mu_{\shape}$ and covariance matrices $\Sigma_{\pose},\Sigma_{\shape}$. The loss is given as the log likelihood of a given shape or pose vector under these distributions, which corresponds to the Mahalanobis distance between the predicted parameters and their corresponding means:
\begin{align}
    \L{pose}(\pose; \mu_{\pose}, \Sigma_{\pose}) &= (\pose - \mu_{\pose})^T \Sigma_{\pose}^{-1} (\pose - \mu_{\pose})\\
    \L{shape}(\shape; \mu_{\shape}, \Sigma_{\shape}) &= (\shape - \mu_{\shape})^T \Sigma_{\shape}^{-1} (\shape - \mu_{\shape})
\end{align}
Unlike previous work, we find there is no need to use a loss to penalize pose parameters if they exceed manually specified joint angle limits. We suspect our network learns this regularization naturally because of our large dataset.

%our network this is a positive side-effect of training a network on a large dataset rather than optimizing independently to single images, as the network can learn natural regularisation that discourages infeasible joint configurations.

%this is since the network is able to use the plentiful training examples to learn its own prior.

\vspace{-1em}
\subsection{Learning a multi-modal shape prior.}

% Using a unimodal prior tends to result in predictions which look relatively similar in shape. To promote diversity among predicted 3D dog shapes, our method extends the formulation above to incorporate a mixture of Gaussians prior. We represent the mixture as a set of $M$ Gaussians, whose means are initialized by drawing samples from our existing prior:

% \begin{align}
%     \mu_{\shape}^{m} &\sim N(\mu_{\shape}, \Sigma_{\shape}) \\
%     \Sigma_{\shape}^{m} &:= \Sigma_{\shape}
% \end{align}

% We assign each training image $i$ with a set of mixture weights $\{w_{i}^{1}, \dots w_{i}^{M}\}$, where initially $w_{i}^{m} := \frac{1}{M}.$

% We can then apply the following mixture shape loss:

% \begin{equation}
%     L_{mixture}=\sum_{m=1}^M w_{i}^{m}L_{shape}(\shape_{i}, \mu_{\shape}^{m}, \Sigma_{\shape}^{m})
% \end{equation}

% In order to allow our mixture prior to learn ``in-the-loop" from the available training data, we apply expectation maximization every $k$ epochs during training. This step recomputes the means and variances for each mixture component based on the observed shapes in the training set, and updates the per-image mixture weights:

% \begin{align}
%     \mu_{\shape}^{m} :=& \mathrm{E}_{i}[\beta_{i}W_{i}^{m}]\\
%     \Sigma_{\shape}^{m} :=& \mathrm{Cov}_{i}[\beta_{i}W_{i}^{m}, \beta_{i}W_{i}^{m}]\\
%     w_{i}^{m} :=& \frac{L_{shape}(\shape_{i}, \mu_{\shape}^{m}, \Sigma_{\shape}^{m})}{\sum_{m'}^{M}L_{shape}(\shape_{i}, \mu_{\shape}^{m'}, \Sigma_{\shape}^{m'})}
% \end{align}

% \section{Attempt 2}

The previous section introduced a unimodal, multivariate Gaussian shape prior, based on mean $\mu_{\shape}$ and covariance matrix $\Sigma_{\shape}$. However, we find enforcing this prior throughout training tends to result in predictions which appear similar in 3D shape, even when tested on dog images of different breeds. We propose to improve diversity among predicted 3D dog shapes by extending the above formulation to a Mixture of $M$ Gaussians prior.  
The mixture shape loss is then given as:
\begin{align}
    \L{mixture}(\shape_{i}; \mu_{\shape}, \Sigma_{\shape}, \Pi_{\shape})
    % =&
    % \sum_{m=1}^M
    % \Pi_{\shape}^m
    % (\shape_{i} - \mu_{\shape}^{m})^{T} \inv{\Sigma_{\shape}^{m}} (\shape_{i} - \mu_{\shape}^{m})
    % \\
    =&
    \sum_{m=1}^M \Pi_{\shape}^{m}\L{shape}(\shape_{i}; \mu_{\shape}^{m}, \Sigma_{\shape}^{m})
\end{align}
Where $\mu_{\shape}^{m}$, $\Sigma_{\shape}^{m}$ and $\Pi_{\shape}^{m}$ 
are the mean, covariance and mixture weight respectively for Gaussian component 
$m$. For each component the mean is sampled from our existing unimodal prior and the covariance is set equal to the unimodal prior i.e. $\Sigma_{\shape}^{m} := \Sigma_{\shape}$. All mixture weights are initially set to $\frac{1}{M}$.

Each training image $i$ is assigned a set of latent variables $\{w_{i}^{1}, \dots w_{i}^{M}\}$ encoding the probability of the dog shape in image~$i$ being generated by component~$m$. 

\vspace{-1em}
\subsection{Expectation Maximization in the loop}

As previously discussed, our initial shape prior is obtained from artist data which we find is unrepresentative of the diverse shapes present in our real dog dataset. We address this by proposing to recover the latent variables $w_{i}^{m}$ and parameters ($\mu_{\shape}^{m}$, $\Sigma_{\shape}^{m}$ and $\Pi_{\shape}^{m}$) of our 3D shape prior by learning from monocular images of in-the-wild dogs and their 2D training labels in our training dataset.

We achieve this using Expectation Maximization (EM), which regularly updates the means and variances for each mixture component and per-image mixture weights based on the observed shapes in the training set. While training our 3D reconstruction network, we progressively update our shape mixture model with an alternating `E' step and `M' step described below:

\subsubsection{The `E' Step.}
The `E' step computes the expected value of the latent variables~$w_{i}^{m}$ 
assuming fixed $(\mu_{\shape}^{m}, \Sigma_{\shape}^{m}, \Pi_{\shape}^{m})$ for all $i \in \{1,\dots,N\}, m \in \{1,\dots,M\}$.

The update equation for an image $i$ with latest shape prediction $\shape_{i}$ 
and cluster $m$ with parameters $(\mu_{\shape}^{m}, \Sigma_{\shape}^{m}, \Pi_{\shape}^{m})$ 
is given as:
% distance between the latest shape prediction $\shape_{i}$ and the cluster $(\mu_{\shape}^{m}, \Sigma_{\shape}^{m})$

% \begin{align}
%     w_{i}^{m} 
%     :=& 
%     \frac{
%         (\shape_{i} - \mu_{\shape}^{m})^{T} \inv{\Sigma_{\shape}^{m}} (\shape_{i} - \mu_{\shape}^{m})
%     }
%     {
%         \sum_{m'}^{M}
%         (\shape_{i} - \mu_{\shape}^{m'})^{T} \inv{\Sigma_{\shape}^{m'}} (\shape_{i} - \mu_{\shape}^{m'})
%     }
%     \\
%     :=&
%     \frac{
%         L_{shape}(\shape_{i}, \mu_{\shape}^{m}, \Sigma_{\shape}^{m})
%     }
%     {
%         \sum_{m'}^{M}L_{shape}(\shape_{i}, \mu_{\shape}^{m'}, \Sigma_{\shape}^{m'})
%     } 
% \end{align}

\begin{align}
    w_{i}^{m} 
    :=& 
    \frac{
        \mathcal{N}(\shape_{i} | \mu_{\shape}^{m},\Sigma_{\shape}^{m})\Pi_{\shape}^{m}
    }
    {
        \sum_{m'}^{M}
        \mathcal{N}(\shape_{i} | \mu_{\shape}^{m'},\Sigma_{\shape}^{m'})\Pi_{\shape}^{m'}
    }
    % \\
    % :=&
    % \frac{
    %     L_{shape}(\shape_{i}, \mu_{\shape}^{m}, \Sigma_{\shape}^{m})
    % }
    % {
    %     \sum_{m'}^{M}L_{shape}(\shape_{i}, \mu_{\shape}^{m'}, \Sigma_{\shape}^{m'})
    % } 
\end{align}

\subsubsection{The `M' Step.}
The `M' step computes new values for $(\mu_{\shape}^{m}, \Sigma_{\shape}^{m}, \Pi_{\shape}^{m})$, assuming fixed $w_{i}^{m}$ for all $i \in \{1,\dots,N\}, m \in \{1,\dots,M\}$.

The update equations are given as follows:

% \begin{align}
%     \mu_{\shape}^{m} :=& 
%     \frac{
%         \sum_{i}w_{i}^{m}\shape_{i}
%     }
%     {
%         \sum_{i}w_{i}^{m}
%     }
%     \\
%     \Sigma_{\shape}^{m} :=& 
%     \frac{
%         \sum_{i}w_{i}^{m}
%         (\shape_{i} - \Sigma_{\shape}^{m})
%         (\shape_{i} - \Sigma_{\shape}^{m})^{T}
%     }
%     {
%         \sum_{i}w_{i}^{m}
%     }
%     \\
%     \Pi_{\shape}^{m} :=& 
%     \frac{1}{N}\sum_{i}{w_{i}^{m}}.
% \end{align}

\begin{equation}
    \mu_{\shape}^{m} := 
    \frac{
        \sum_{i}w_{i}^{m}\shape_{i}
    }
    {
        \sum_{i}w_{i}^{m}
    }
    \quad
    \Sigma_{\shape}^{m} :=
    \frac{
        \sum_{i}w_{i}^{m}
        (\shape_{i} - \Sigma_{\shape}^{m})
        (\shape_{i} - \Sigma_{\shape}^{m})^{T}
    }
    {
        \sum_{i}w_{i}^{m}
    }
    \quad
    \Pi_{\shape}^{m} :=
    \frac{1}{N}\sum_{i}{w_{i}^{m}}
\end{equation}

\section{Experiments}

% As with other methods, we observe that introducing a silhouette term into the network loss before the model's pose has been somewhat solved can result in unsatisfactory local minima. We overcome this by using a pre-training stage with the following loss terms:

In this section we compare our method to competitive baselines. We begin by describing our new large-scale dataset of annotated dog images, followed by a quantitative and qualitative evaluation.

\vspace{-1em}
\subsection{StanfordExtra: A new large-scale dog dataset with 2D keypoint and silhouette annotations}

\begin{figure*}[h]
    \centering
    \includegraphics[height=0.1775\textheight]{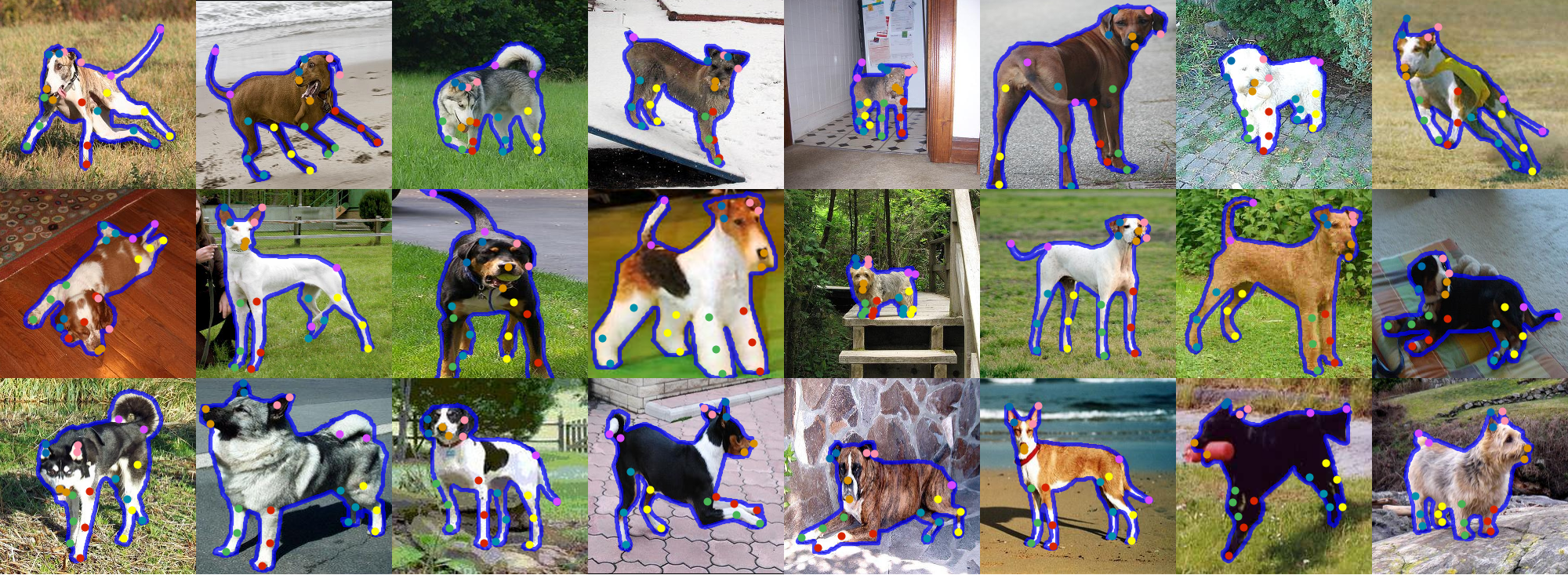}
    \includegraphics[height=0.1775\textheight]{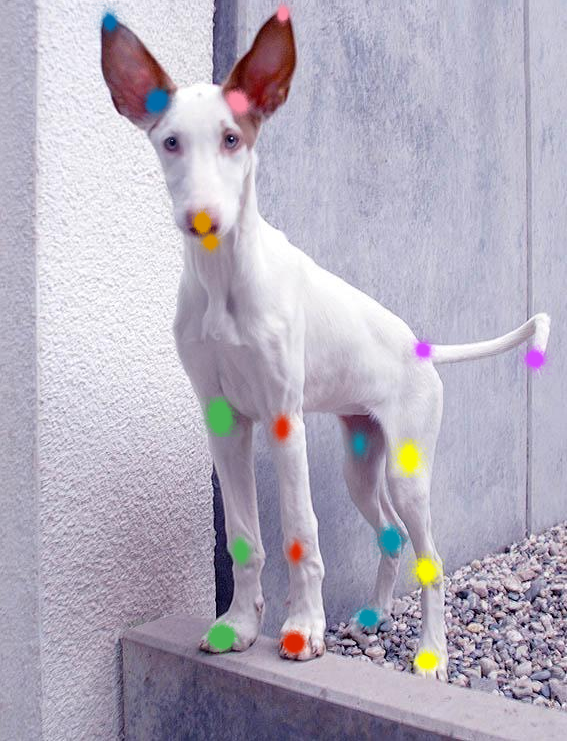}
    \caption{\textbf{StanfordExtra example images}. \emph{Left}: outlined segmentations and labelled keypoints for 24 representative images. \emph{Right}: heatmap of deviation of worker submitted results from mean for each submission.}
    \label{fig:dataset}
\end{figure*}

In order to evaluate our method, we introduce \emph{StanfordExtra}: a new large-scale dataset with annotated 2D keypoints and binary segmentation masks for dogs. We opted to take source images from the existing Stanford Dog Dataset~\cite{StanfordDogs}, which consists of 20,580 dog images taken ``in the wild" and covers 120 dog breeds. The dataset contains vast shape and pose variation between dogs, as well as nuisance factors such as self/environmental occlusion, interaction with humans/other animals and partial views. Figure~\ref{fig:dataset} (left) shows samples from the new dataset.

We used Amazon Mechanical Turk to collect a binary silhouette mask and 20 keypoints per image: 3 per leg (knee, ankle, toe), 2 per ear (base, tip), 2 per tail (base, tip), 2 per face (nose and jaw). We can approximate the difficulty of the dataset by analysing the variance between 3 annotators at both the joint labelling and silhouette task. Figure~\ref{fig:dataset} (right) illustrates typical per-joint variance in joint labelling. Further details of the data curation procedure are left to the supplementary materials. 

\vspace{-1em}
\subsection{Evaluation protocol}

Our evaluation is based on our new StanfordExtra dataset. In line with other ``in-the-wild'' 3D reconstruction methods tackling articulated subjects~\cite{kolotouros2019learning,kolotouros19convolutional}, we filter images from the original dataset of 20,580 for which the majority of dog keypoints are invisible. We consider these images unsuitable for our full-body dog reconstruction task. We also remove images for which the consistency in keypoint/silhouette segmentations between the 3 annotators is below a set threshold. This leaves us with 8,476 images which we divide per-breed into an 80\%/20\% train and test split.

We consider two primary evaluation metrics. IoU is the intersection-over-union of the projected model silhouette compared to the ground truth annotation and indicates the quality of the reconstructed 3D shape. Percentage of Correct Keypoints (PCK) computes the percentage of joints which are within a normalized distance (based on square root of 2D silhouette area) to the ground truth locations, and evaluates the quality of reconstructed 3D pose. We also produce PCK results on various joint groups (legs, tail, ears, face) to compare the reconstruction accuracy for different parts of the dog model.

\vspace{-1em}
\subsection{Training procedure}

We train our model in two stages. The first omits the silhouette loss which we find can lead the network to unsatisfactory local minima if applied too early. With the silhouette loss turned off, we find it satisfactory to use the simple unimodal prior (and without EM) for this preliminary stage since there is no loss to specifically encourage a strong shape alignment. After this, we introduce the silhouette loss, the mixture prior and begin applying the expectation maximization updates over $M=10$ clusters. We train the first stage for 250 epochs, the second stage for 150 and apply the EM step every 50 epochs. All losses are weighted, as described in the supplementary. The entire training procedure takes 96 hours on a single P100 GPU.

\vspace{-1em}
\subsection{Comparison to baselines}

We first compare our method to various baseline methods. 3D Menagerie (3D-M)~\cite{DBLP:journals/corr/ZuffiKJB16} is an approach which fits the 3D SMAL model using per-image energy minimization. Creatures Great and SMAL (CGAS)~\cite{biggs2018creatures} is a three-stage method, which employs a joint predictor on silhouette renderings from synthetic 3D dogs, applies a genetic algorithm to clean predictions, and finally applies the SMAL optimizer to produce the 3D mesh.

At test-time both 3D-M and CGAS rely on manually-provided segementation masks, and 3D-M also relies on hand-clicked keypoints. In order to produce a fair comparison, we produce a set of \emph{predicted} keypoints for StanfordExtra by training the Stacked Hourglass Network~\cite{newell2016stacked} with 8 stacks and 1 block, and \emph{predicted} segmentation masks using DeepLab v3+~\cite{journals/corr/ChenPK0Y16}. The Stacked Hourglass Network achieves 71.4\% PCK score, DeepLab v3+ achieves 83.4\% IoU score and the CGAS joint predictor achieves 41.8\% PCK score. 

%All methods are trained from scratch and evaluated on our Stanford Dog validation set.

% \input{eccv2020kit/tab_othernetworks}
% \input{tab_othernetworks}

Table~\ref{tab:baselinesfix} and Figure~\ref{fig:comparison_sup} show the comparison against competitive methods. For full examination, we additionally provide results for 3D-M and CGAS in the scenario that ground-truth keypoints and/or segmentations are available at test time. 

The results show our end-to-end method outperforms the competitors when they are provided with predicted keypoints/segmentations (white rows). Our method therefore achieves a new state-of-the-art on this 3D reconstruction task. In addition, we show our method achieves improved average IoU/PCK scores than competitive methods, even when they are provided ground truth annotations at test time (grey rows). We also demonstrate wider applicability of two contributions from our work (scale parameters and improved prior) by showing improved performance of the 3D-M method when these are incorporated. Finally, our model's test-time speed is significantly faster than the competitors as it does not require an optimizer.

\begin{table}[]
{
    \small
    \centering
    \begin{tabular}{@{}lcccccccc@{}}
    \toprule
    \multicolumn{1}{l}{Method} & 
    \multicolumn{1}{c}{Kps} & 
    \multicolumn{1}{c}{Seg} & 
    \multicolumn{1}{c}{IoU} & 
    \multicolumn{5}{c}{PCK @ 0.15} \\
    \multicolumn{4}{c}{} &
    \multicolumn{1}{c}{Avg} &
    \multicolumn{1}{c}{Legs} &
    \multicolumn{1}{c}{Tail} &
    \multicolumn{1}{c}{Ears} &
    \multicolumn{1}{c}{Face} \\
    \midrule
    \rowcolor{comp} 3D-M~\cite{DBLP:journals/corr/ZuffiKJB16} & Pred & Pred & 69.9 & 69.7 & 68.3 & 68.0 & 57.8 & 93.7 \\
    \rowcolor{notcomp} 3D-M & GT & GT & 71.0 & 75.6 & 74.2 & 89.5 & 60.7 & 98.6 \\
    \rowcolor{notcomp} 3D-M & GT & Pred & 70.7 & 75.5 & 74.1 & 88.1 & 60.2 & 98.7 \\
    \rowcolor{notcomp} 3D-M & Pred & GT & 70.5 & 70.3 & 69.0 & 69.4 & 58.5 & 94.0 \\
    \hline
    \rowcolor{comp} CGAS~\cite{biggs2018creatures} & CGAS & Pred & 63.5 & 28.6 & 30.7 & 34.5 & 25.9 & 24.1 \\
    \rowcolor{notcomp} CGAS & CGAS & GT & 64.2 & 28.2 & 30.1 & 33.4 & 26.3 & 24.5 \\
    \hline
    \rowcolor{comp} 3D-M + scaling & Pred & Pred & 70.4 & 70.9 & 69.8 & 66.9 & 59.7 & 94.0 \\
    \rowcolor{comp} 3D-M + scaling + EM prior & Pred & Pred & 71.8 & 73.4 & 72.5 & \textbf{70.3} & 62.6 & \textbf{94.1} \\
    \hline
    \rowcolor{comp} \textbf{Ours} & --- & --- & \textbf{74.2} & \textbf{78.8} & \textbf{76.4} & 63.9 & \textbf{78.1} & 92.1 \\
    \bottomrule 
    \end{tabular}
    \vspace{1em}
    \caption[]{\label{tab:baselinesfix}\textbf{Baseline comparisons.\footnotemark[4]} PCK and silhouette IOU scores are shown for SOTA methods under varying conditions. Directly comparable baseline methods (requiring only an input image) are highlighted. \emph{Pred} keypoints generated with Hourglass-Net~\cite{newell2016stacked} and segmentations with DeepLab v3+~\cite{journals/corr/ChenPK0Y16}. 3D-M/CGAS are also analysed when they have access to ground-truth keypoints and/or segmentation masks. We also analyse adding this paper's innovations (scale parameters and EM prior) to 3D-M~\cite{DBLP:journals/corr/ZuffiKJB16}.}
}
\end{table}
\newcolumntype{?}{!{\vrule width 1pt}}

\newcommand\sfac{0.082}
\newcommand\spacercomp{4mm}
\begin{figure*}[ht!]
    \centering
    \setkeys{Gin}{width=\linewidth}
    \renewcommand\tabularxcolumn[1]{>{\Centering}m{\sfac\linewidth}} % set all columns to be centered v & hwise, with a fixed length
    \begin{tabularx}{\textwidth}{c*{5}{X}@{\hspace{\spacercomp}}*{5}{X}c}
       \textbf{Ours} &
      \includegraphics{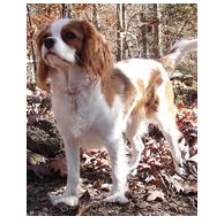} &
      \includegraphics{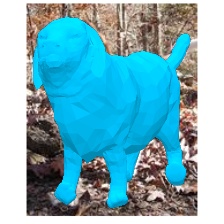} &
      \includegraphics{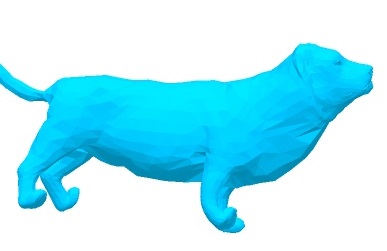} &
      \includegraphics{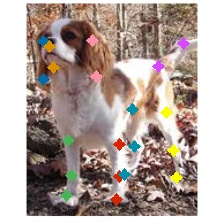} &
      \includegraphics{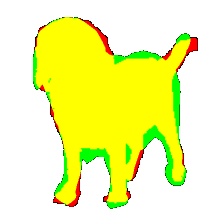} &
      \includegraphics{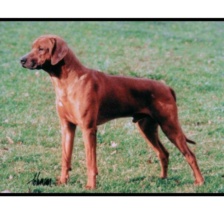} &
      \includegraphics{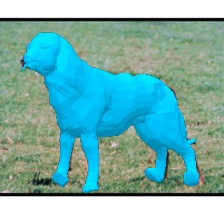} &
      \includegraphics{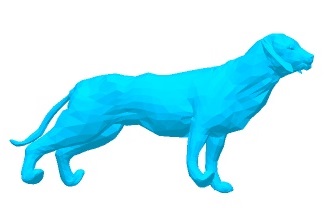} &
      \includegraphics{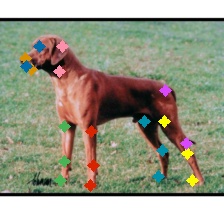} &
      \includegraphics{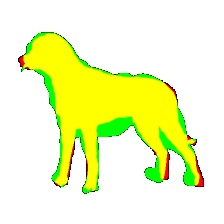} \\

       3D-M &
      \includegraphics{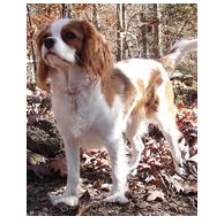} &
      \includegraphics{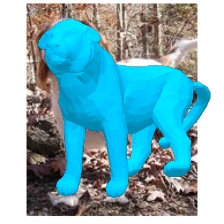} &
      \includegraphics{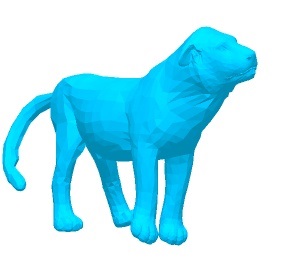} &
      \includegraphics{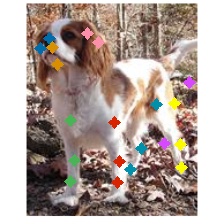} &
      \includegraphics{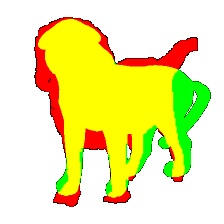} &
      %\hspace{\spacercomp}
      \includegraphics{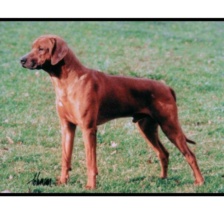} &
      \includegraphics{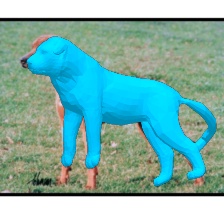} &
      \includegraphics{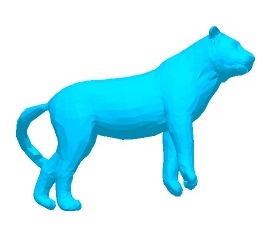} &
      \includegraphics{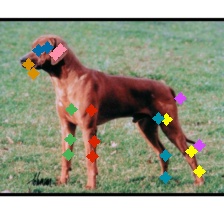} &
      \includegraphics{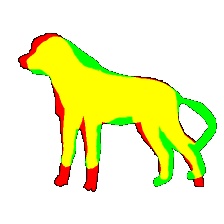} \\

       CGAS &
      \includegraphics{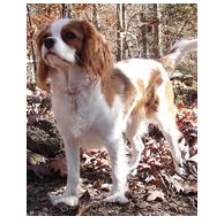} &
      \includegraphics{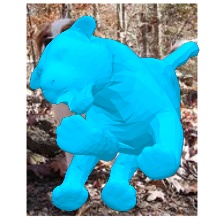} &
      \includegraphics{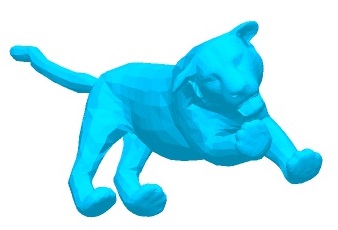} &
      \includegraphics{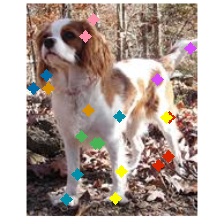} &
      \includegraphics{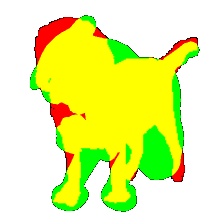} &
      %\hspace{\spacercomp}
      \includegraphics{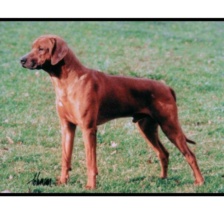} &
      \includegraphics{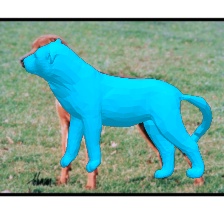} &
      \includegraphics{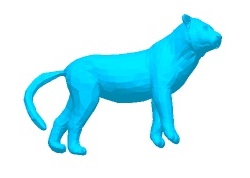} &
      \includegraphics{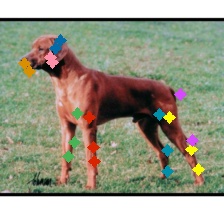} &
      \includegraphics{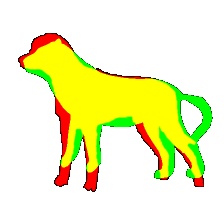} \\

      \textbf{Ours} &
      \includegraphics{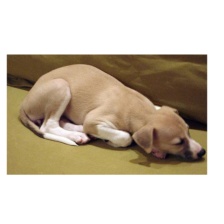} &
      \includegraphics{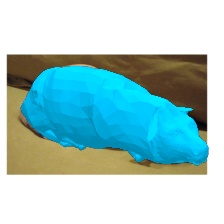} &
      \includegraphics{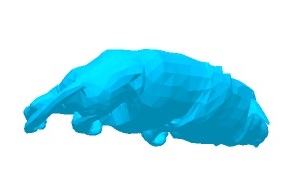} &
      \includegraphics{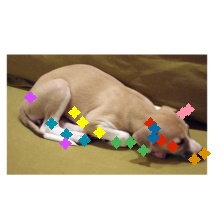} &
      \includegraphics{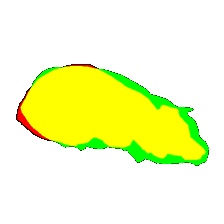} &
      
        \includegraphics{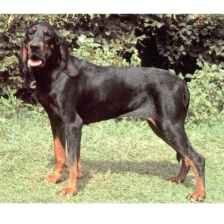} &
      \includegraphics{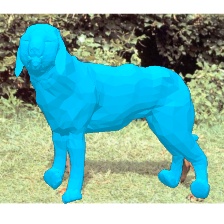} &
      \includegraphics{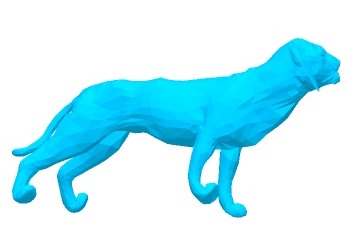} &
      \includegraphics{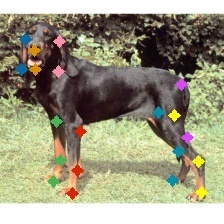} &
      \includegraphics{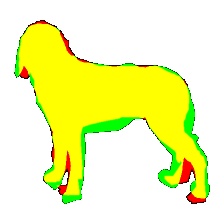} \\

        3D-M & 
      \includegraphics{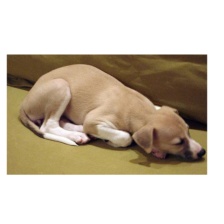} &
      \includegraphics{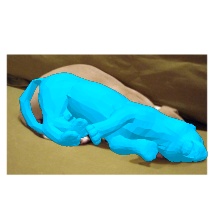} &
      \includegraphics{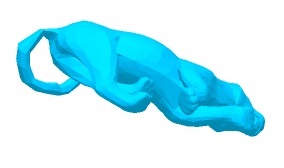} &
      \includegraphics{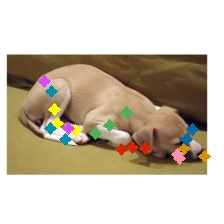} &
      \includegraphics{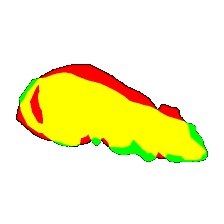} &
      
      \includegraphics{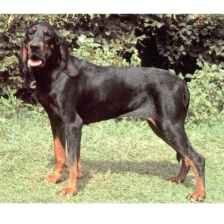}&
      \includegraphics{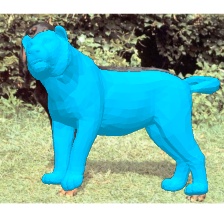}&
      \includegraphics{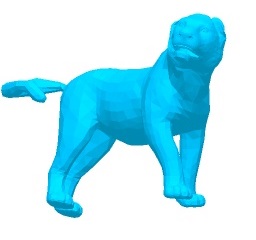}&
      \includegraphics{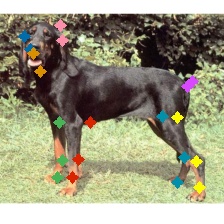}&
      \includegraphics{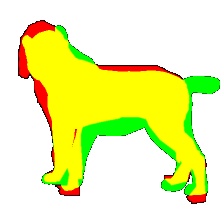}
      \\
      
      CGAS &
      \includegraphics{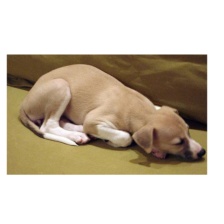} &
      \includegraphics{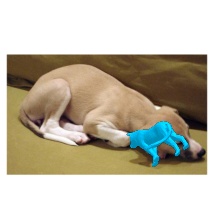} &
      \includegraphics{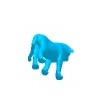} &
      \includegraphics{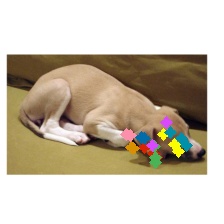} &
      \includegraphics{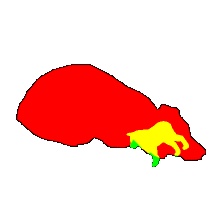} &
      
    \includegraphics{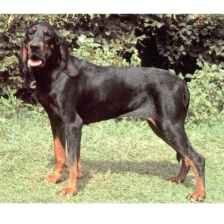}&
      \includegraphics{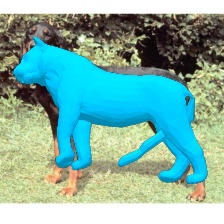}&
      \includegraphics{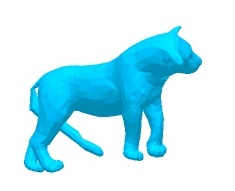}&
      \includegraphics{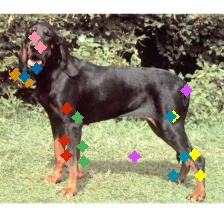}&
      \includegraphics{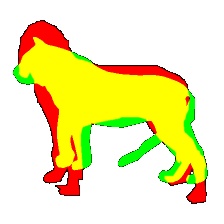}
      \\

      & (a) & (b) & (c) & (d) & (e) & 
      %\hspace{\spacercomp} 
      (a) & (b) & (c) & (d) & (e) \\
    \end{tabularx}
    \caption{%
    \textbf{Qualitiative comparison to SOTA.} 
    Row 1: \textbf{Ours}, 
    Row 2: 3D-M~\cite{DBLP:journals/corr/ZuffiKJB16}, 
    Row 3: CGAS~\cite{biggs2018creatures}. 
    (a) input image, (b) predicted 3D mesh, (c) canonical view 3D mesh, 
    (d) reprojected model joints and (e) silhouette reprojection error. 
    }
    \label{fig:comparison_sup}
\end{figure*}

%xxx: Note that CGAS does badly as it can't clean up using video
\vspace{-1em}
\subsection{Generalization to unseen dataset}

Table~\ref{tab:animalposefix} shows an experiment to compare how well our model generalizes to a new data domain. We test our model against the 3D-M~\cite{DBLP:journals/corr/ZuffiKJB16} method (using predicted keypoints and segmentations as above for fairness) on the recent Animal Pose dataset~\cite{animalpose}. The data preparation process is the same as for StanfordExtra and no fine-tuning was used for either method. We achieve good results in this unseen domain and still improve over the 3D-M optimizer.

\begin{table}
    \parbox{.45\linewidth}{
        % \small
        \centering
        \begin{tabular}{@{}lcccccc@{}}
        \toprule
        \multicolumn{1}{l}{Method} & 
        \multicolumn{1}{c}{IoU} & 
        \multicolumn{5}{c}{PCK @ 0.15} \\
        \multicolumn{2}{c}{} &
        \multicolumn{1}{c}{Avg} &
        \multicolumn{1}{c}{Legs} &
        \multicolumn{1}{c}{Tail} &
        \multicolumn{1}{c}{Ears} &
        \multicolumn{1}{c}{Face} \\
        \midrule
        3D-M~\cite{DBLP:journals/corr/ZuffiKJB16} & 64.9 & 59.2 & 55.7 & 56.9 & 61.3 & \textbf{86.7} \\
        % \hline
        \textbf{Ours} & \textbf{67.5} & \textbf{67.6} & \textbf{60.4} & \textbf{62.7} & \textbf{86.0} & \textbf{86.7} \\
        \bottomrule
        \multicolumn{7}{c}{} \\
        \multicolumn{7}{c}{}
        \end{tabular}
        \vspace{1em}
        \caption[]{
            \label{tab:animalposefix}
            \textbf{Animal Pose dataset~\cite{animalpose} results\footnotemark[4]}. Evaluation on recent Animal Pose dataset with no fine-tuning to our method nor joint/silhouette predictors used for 3D-M.
        }
    }
    \hfill
    \parbox{.45\linewidth}{
        % \small
        \centering
        \begin{tabular}{@{}lcccccc@{}}
        \toprule
        \multicolumn{1}{l}{Method} & 
        \multicolumn{1}{c}{IoU} & 
        \multicolumn{5}{c}{PCK @ 0.1} \\
        \multicolumn{2}{c}{} &
        \multicolumn{1}{c}{Avg} &
        \multicolumn{1}{c}{Legs} &
        \multicolumn{1}{c}{Tail} &
        \multicolumn{1}{c}{Ears} &
        \multicolumn{1}{c}{Face} \\
        \midrule
        \textbf{Ours} & \textbf{74.2} & \textbf{63.7} & \textbf{59.5} & \textbf{48.1} & 60.1 & 88.0 \\
        $-$EM & 68.7 & 63.2 & 58.8 & 44.5 & \textbf{62.6} & 87.6 \\
        $-$MoG & 69.0 & 63.1 & \textbf{59.5} & 40.0 & 60.0 & \textbf{89.5} \\
        $-$Scale & 68.3 & 60.1 & 58.2 & 45.2 & 50.5 & 88.3 \\
        \bottomrule 
        \end{tabular}
        \vspace{1em}
        \caption[]{
            \label{tab:ablationfix}\textbf{Ablation study.\footnotemark[4]} Evaluation with the following contributions removed: (a) EM updates, (b) Mixture Shape Prior, (c) SMBLD scale parameters.
        }
    }
\end{table}

\vspace{-1em}
\subsection{Ablation study}

We also produce a study in which we ablate individual components of our method and examine the effect on the PCK/IoU performance. We evaluate three variants: (1) \textbf{Ours w/o EM} that omits EM updates, (2) \textbf{Ours w/o MoG} which replaces our mixture shape prior with a unimodal prior, (3)~\textbf{Ours w/o Scale} which removes the scale parameters. 

The results in Table~\ref{tab:ablationfix} indicate that each individual component has a positive impact on the overall method performance. In particular, it can be seen that the inclusion of the EM and Mixture of Gaussians prior leads to an improvement in IoU, suggesting that the shape prior refinements steps help the model accurately fit the exact dog shape. Interestingly, we notice that adding the Mixture of Gaussians prior but omitting EM steps slightly hinders performance, perhaps due to an sub-optimal initialization for the $M$ clusters. However, we find adding EM updates to the Mixture of Gaussian model improves all metrics except the ear keypoint accuracy. We observe the error here is caused by the our shape prior learning slightly imprecise shapes for dogs with extremely ``floppy'' ears. Although there is good silhouette coverage for these regions, the fact our model has only a single articulation point per ear causes a lack of flexibility that results in occasionally misplaced ear tips for these instances. This could be improved in future work by adding additional model joints to the ear. Finally, we find the increased model flexibility afforded by the SMBLD scale parameters have a positive effect on IoU/PCK scores. 

% \input{eccv2020kit/tab_ablation}

% We are able to use the dataset to examine which dog parts are the most challenging to position. 
% \input{eccv2020kit/fig_jointspreads}
% \anote{TODO: PCK tables and errors visualized on 3D dog.}
% \paragraph{Analysis over breeds}
% A significant benefit of our dog dataset is that the supplied breed labels allows for reconstruction performance to be evaluated over particular breeds. \anote{Figure} ranks the breeds by error.
% \input{eccv2020kit/tab_breed}

\subsection{Qualitative evaluation}

Figure~\ref{fig:comparison_sup} shows a range of example system outputs when tested on range of StanfordExtra and Animal Pose~\cite{animalpose} dogs with varying pose and shape and in challenging conditions. Note that only StanfordExtra is used for training.

% \input{fig_comparison}

% \input{fig_qualresults}

% \input{fig_qual_results_animal_pose}
% \section{Failure Cases}

\footnotetext[4]{PCK results in tables have been updated to match definitions of Yang and Ramanan~\cite{yang2013articulated} normalized by 2D silhouette area. Please see original tables and further details in the appendix.}

\section{Conclusions}
This paper presents an end-to-end method for automatic, monocular 3D dog reconstruction. We achieve this using only weak 2D supervision, provided by our novel StanfordExtra dataset. Further, we show we can learn a more detailed shape prior by tuning a gaussian mixture during model training and this leads to improved reconstructions. We also show our method improves over competitive baselines, even when they are given access to ground truth data at test time.

Future work should involve tackling some failure cases of our system, for example handling multiple overlapping dogs or dealing with heavy motion blur. Other areas for research include extending our EM formulation to handle video input to take advantage of multi-view shape constraints, and transferring knowledge accumulated through training on StanfordExtra dogs to other species.

\section{Acknowlegements}
The authors would like to thank the GSK AI team for providing access to their GPU cluster, Michael Sutcliffe, Thomas Roddick, Matthew Allen and Peter Fisher for useful technical discussions, and the GSK TDI group for project sponsorship. 

\newcolumntype{?}{!{\vrule width 1pt}}

\newcolumntype{M}[1]{>{\centering\arraybackslash}m{#1}}
\newcommand\scalefactorqual{0.085}
\newcommand\spacerqual{3mm}
\begin{figure*}[t!]
    \centering
    \setkeys{Gin}{width=\linewidth}
    %M{40pt}
    \renewcommand\tabularxcolumn[1]{>{\Centering}m{\scalefactorqual\linewidth}} % set all columns to be centered v & hwise, with a fixed width
    \begin{tabularx}{\textwidth}{m{15pt}*{5}{X} @ {\hspace{\spacercomp}}*{5}{X}}%
        %260pt = 8 rows, 
        \multirow{-2}{*}{\rotatebox[origin=c]{90}{$\overbrace{\hspace{293pt}}^{\textrm{\large StanfordExtra}}$}} &
    
        % R1
        \includegraphics{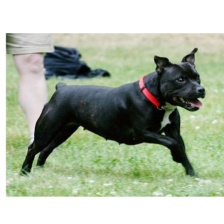} &
        \includegraphics{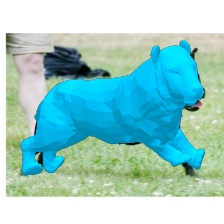} &
        \includegraphics{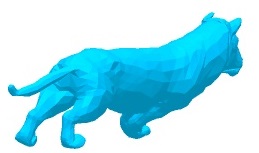} &
        \includegraphics{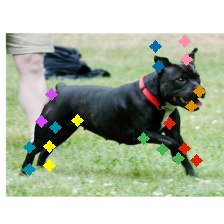} &
        \includegraphics{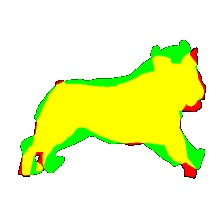} & 
        %\hspace{\spacercomp} 
        \includegraphics{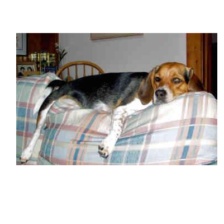} & 
        \includegraphics{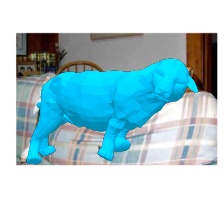} & 
        \includegraphics{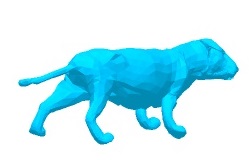} & 
        \includegraphics{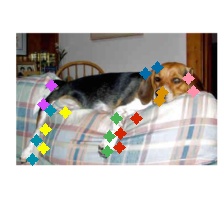} & 
        \includegraphics{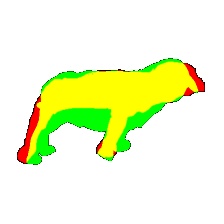} \\ 

        % R2
        &\includegraphics{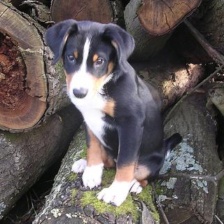} & 
        \includegraphics{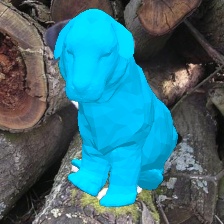} & 
        \includegraphics{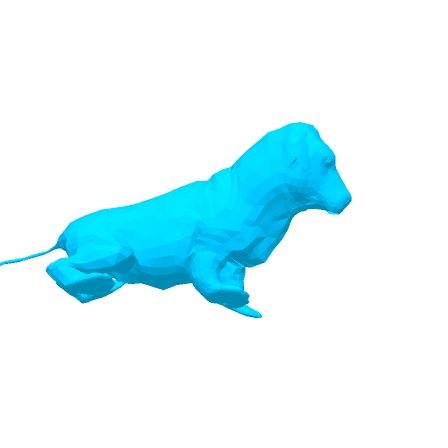} & 
        \includegraphics{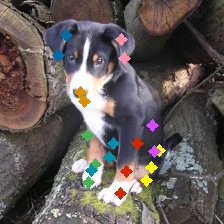} & 
        \includegraphics{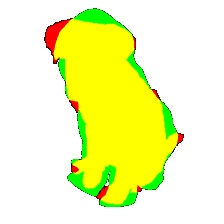} & 
        %\hspace{\spacercomp} 
        \includegraphics{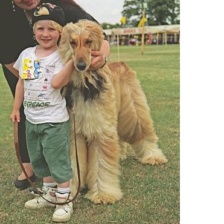} & 
        \includegraphics{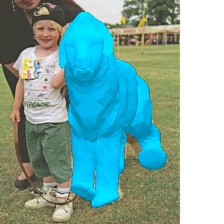} & 
        \includegraphics{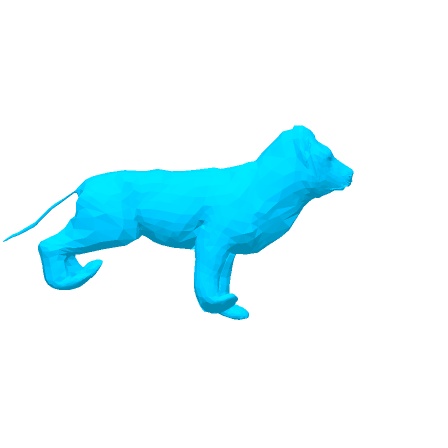} & 
        \includegraphics{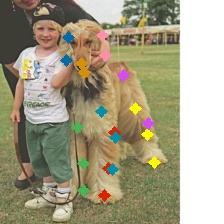} & 
        \includegraphics{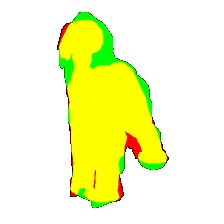} \\ 

        % R3
        &\includegraphics{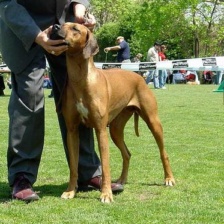} & 
        \includegraphics{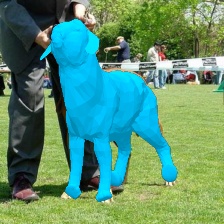} & 
        \includegraphics{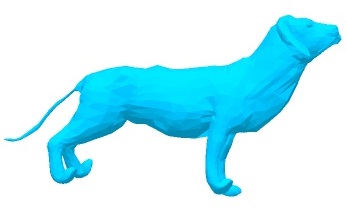} & 
        \includegraphics{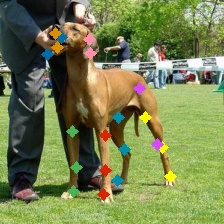} & 
        \includegraphics{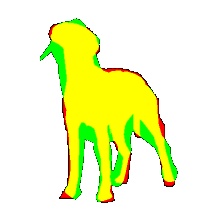} &
        %\hspace{\spacercomp} 
        \includegraphics{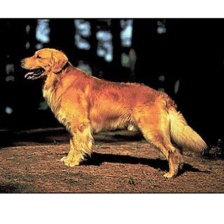} & 
        \includegraphics{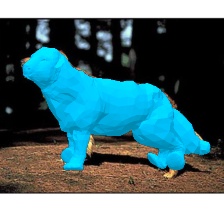} & 
        \includegraphics{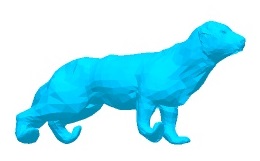} & 
        \includegraphics{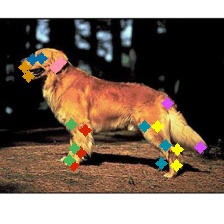} & 
        \includegraphics{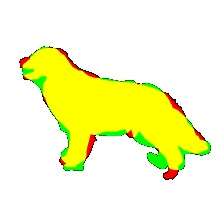} \\ 
        
        % R4
        &\includegraphics{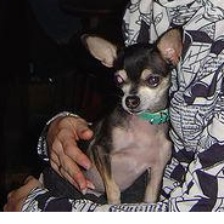} & 
        \includegraphics{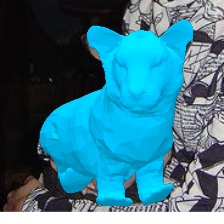} & 
        \includegraphics{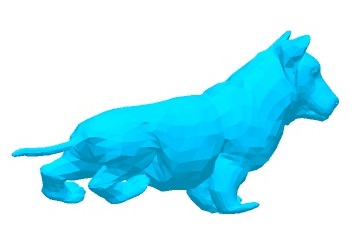} &
        \includegraphics{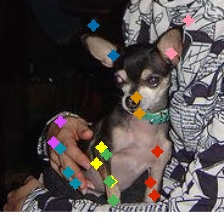} &
        \includegraphics{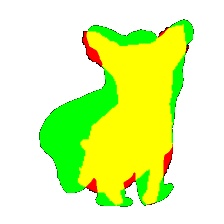} &
        %\hspace{\spacercomp} 
        % \includegraphics{ours_sup/n02091134-whippet/orig/n02091134_16201.jpg} &
        % \includegraphics{ours_sup/n02091134-whippet/fit/n02091134_16201.jpg} &
        % \includegraphics{ours_sup/n02091134-whippet/model/n02091134_16201.jpg} &
        % \includegraphics{ours_sup/n02091134-whippet/joints/n02091134_16201.jpg} &
        % \includegraphics{ours_sup/n02091134-whippet/segs/n02091134_16201.jpg} \\ 
        \includegraphics{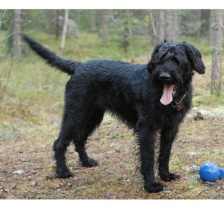} &
        \includegraphics{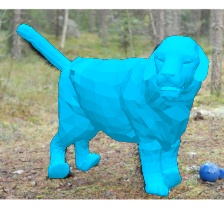} &
        \includegraphics{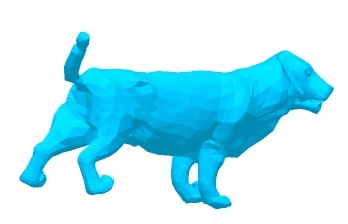} &
        \includegraphics{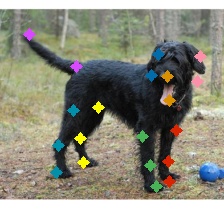} &
        \includegraphics{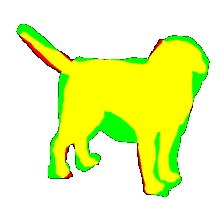} \\
        
        % R5
        &\includegraphics{ours_sup/n02089078-black-and-tan_coonhound/orig/n02089078_877.jpg} &
        \includegraphics{ours_sup/n02089078-black-and-tan_coonhound/fit/n02089078_877.jpg} &
        \includegraphics{ours_sup/n02089078-black-and-tan_coonhound/model/n02089078_877_crop.jpg} &
        \includegraphics{ours_sup/n02089078-black-and-tan_coonhound/joints/n02089078_877.jpg} &
        \includegraphics{ours_sup/n02089078-black-and-tan_coonhound/segs/n02089078_877.jpg} &
        %\hspace{\spacercomp} 
        \includegraphics{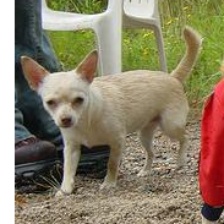} &
        \includegraphics{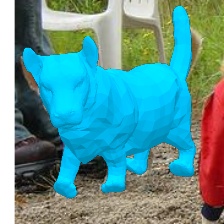} &
        \includegraphics{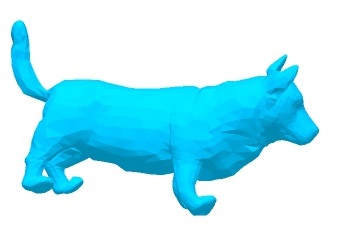} &
        \includegraphics{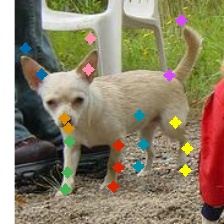} &
        \includegraphics{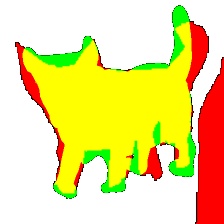} \\ 
        
        % R6
        &\includegraphics{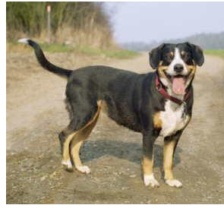} &
        \includegraphics{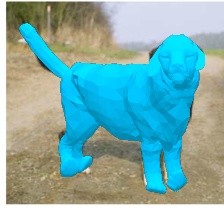} &
        \includegraphics{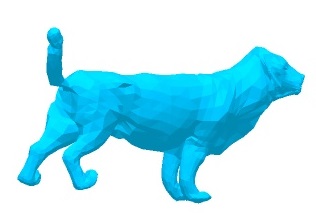} &
        \includegraphics{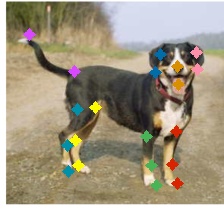} &
        \includegraphics{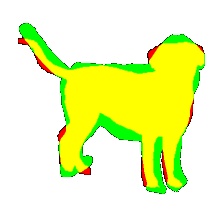} &
        %\hspace{\spacercomp} 
        \includegraphics{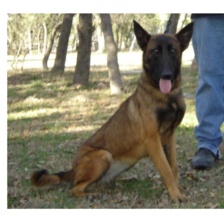} &
        \includegraphics{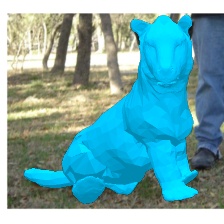} &
        \includegraphics{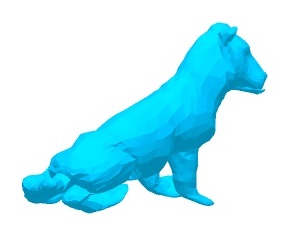} &
        \includegraphics{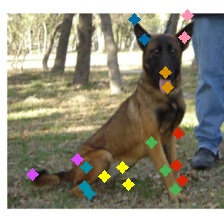} &
        \includegraphics{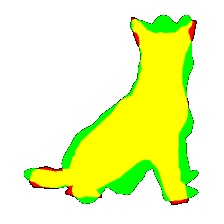} \\ 
        
        % R7
        &\includegraphics{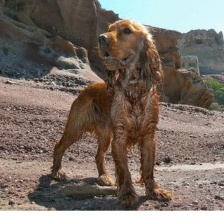} &
        \includegraphics{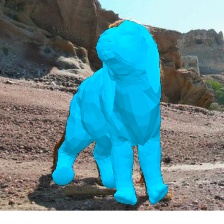} &
        \includegraphics{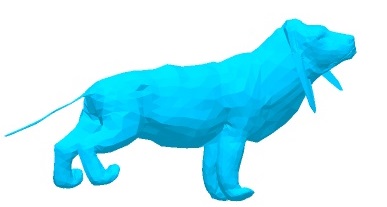} &
        \includegraphics{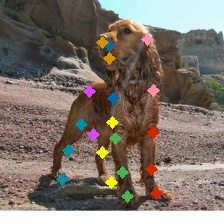} &
        \includegraphics{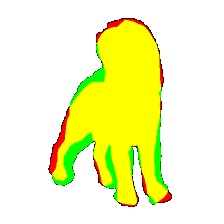} &
        %\hspace{\spacercomp}
        \includegraphics{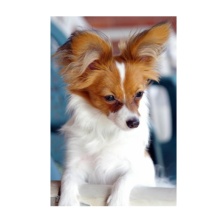} &
        \includegraphics{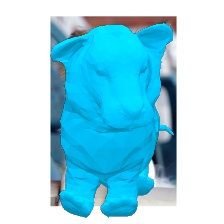} &
        \includegraphics{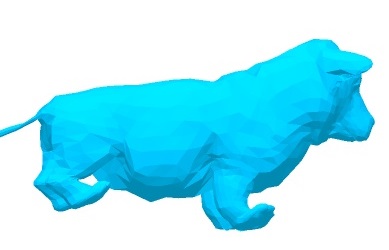} &
        \includegraphics{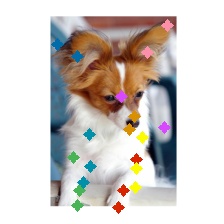} &
        \includegraphics{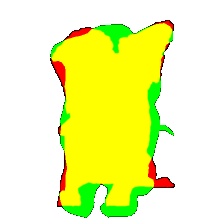} \\
        
        % R8
        % &\includegraphics{ours_sup/n02097130-giant_schnauzer/orig/n02097130_5121.jpg} &
        % \includegraphics{ours_sup/n02097130-giant_schnauzer/fit/n02097130_5121.jpg} &
        % \includegraphics{ours_sup/n02097130-giant_schnauzer/model/n02097130_5121.jpg} &
        % \includegraphics{ours_sup/n02097130-giant_schnauzer/joints/n02097130_5121.jpg} &
        % \includegraphics{ours_sup/n02097130-giant_schnauzer/segs/n02097130_5121.jpg} &
        % \hspace{\spacercomp} 
        % \includegraphics{ours_sup/n02087394-Rhodesian_ridgeback/orig/n02087394_831.jpg} &
        % \includegraphics{ours_sup/n02087394-Rhodesian_ridgeback/fit/n02087394_831.jpg} &
        % \includegraphics{ours_sup/n02087394-Rhodesian_ridgeback/model/n02087394_831.jpg} &
        % \includegraphics{ours_sup/n02087394-Rhodesian_ridgeback/joints/n02087394_831.jpg} &
        % \includegraphics{ours_sup/n02087394-Rhodesian_ridgeback/segs/n02087394_831.jpg} \\

        % R9
        &\includegraphics{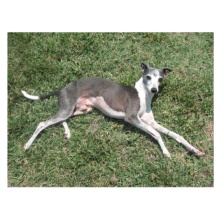} &
        \includegraphics{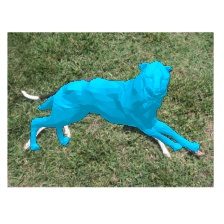} &
        \includegraphics{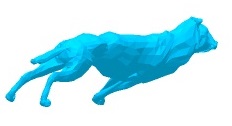} &
        \includegraphics{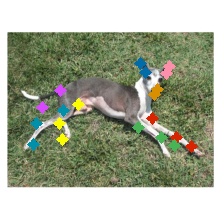} &
        \includegraphics{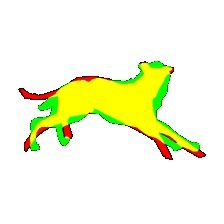} &
        %\hspace{\spacercomp}
        \includegraphics{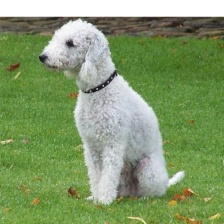} &
        \includegraphics{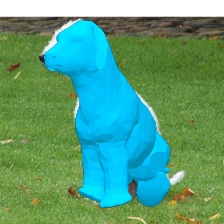} &
        \includegraphics{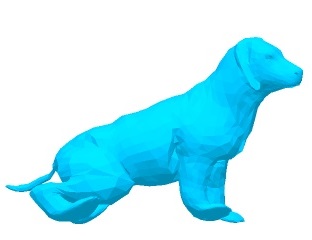} &
        \includegraphics{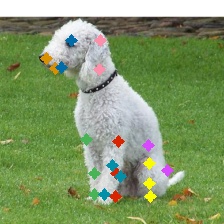} &
        \includegraphics{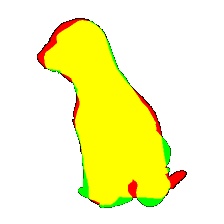} \\

        %R10
        &\includegraphics{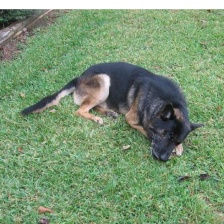} &
        \includegraphics{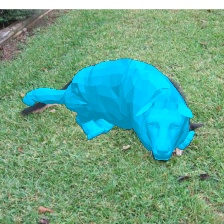} &
        \includegraphics{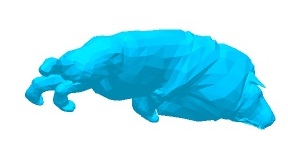} &
        \includegraphics{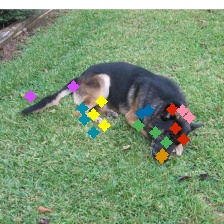} &
        \includegraphics{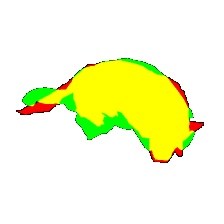} &
        %\hspace{\spacercomp} 
        \includegraphics{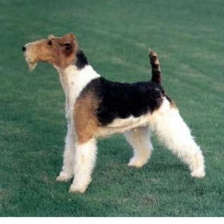} &
        \includegraphics{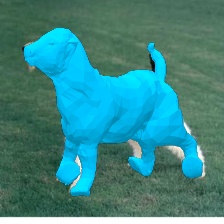} &
        \includegraphics{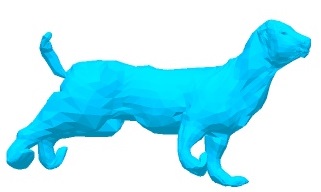} &
        \includegraphics{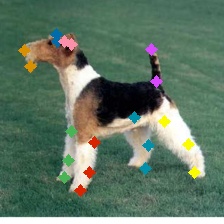} &
        \includegraphics{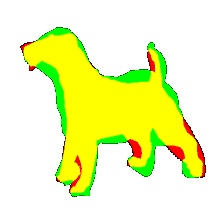} \\

        % R11
        \multirow{-2.1}{*}{\rotatebox[origin=c]{90}{$\overbrace{\hspace{64pt}}^{\textrm{\large Animal Pose}}$}} &
        
        \includegraphics{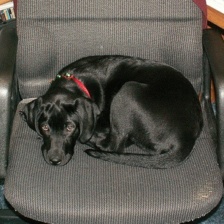} &
        \includegraphics{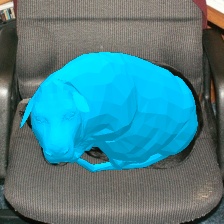} &
        \includegraphics{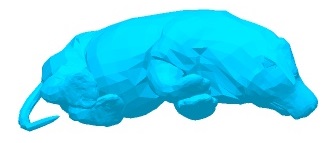} &
        \includegraphics{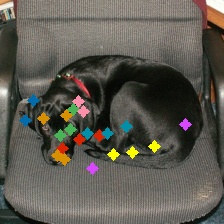} &
        \includegraphics{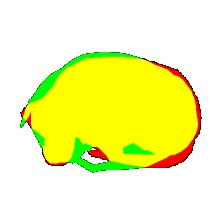} & 
        %\hspace{\spacercomp} 
        \includegraphics{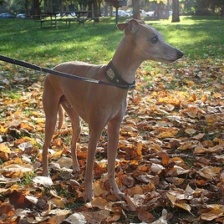} &
        \includegraphics{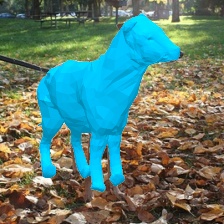} &
        \includegraphics{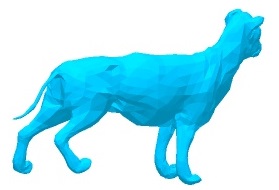} &
        \includegraphics{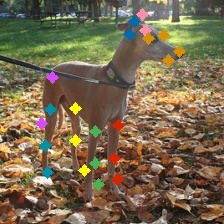} &
        \includegraphics{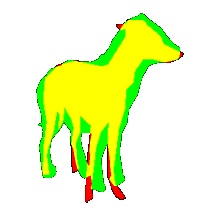} \\ 

        % R12
        &\includegraphics{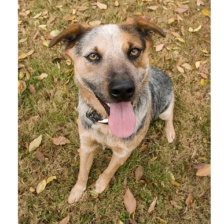} &
        \includegraphics{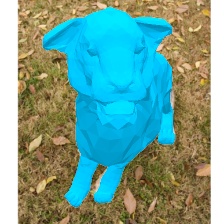} &
        \includegraphics{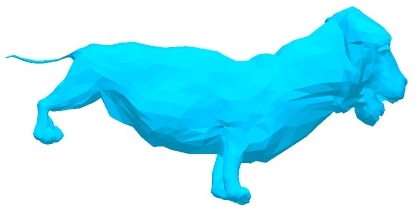} &
        \includegraphics{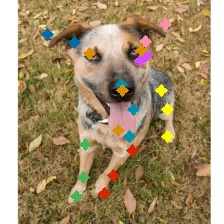} &
        \includegraphics{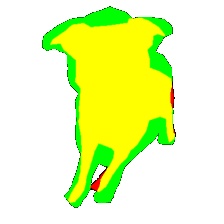} & 
        %\hspace{\spacercomp} 
        \includegraphics{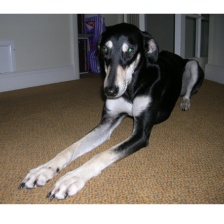} &
        \includegraphics{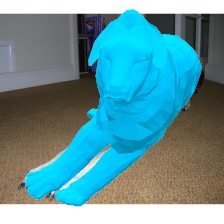} &
        \includegraphics{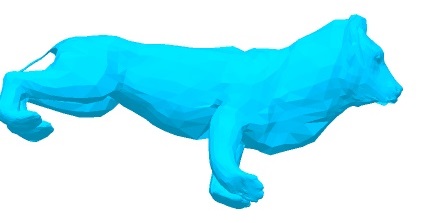} &
        \includegraphics{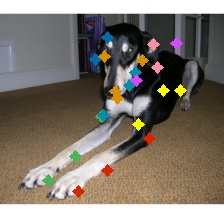} &
        \includegraphics{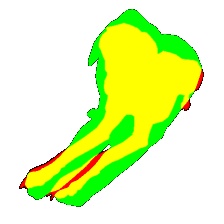} \\ 
        
        & (a) & (b) & (c) & (d) & (e) & 
        %\hspace{\spacercomp} 
        (a) & (b) & (c) & (d) & (e) \\

    \end{tabularx}
    \caption{%
    \textbf{Qualitative results on StanfordExtra and Animal Pose~\cite{animalpose}.} 
        For each sample we show: (a) input image, (b) predicted 3D mesh, 
        (c) canonical view 3D mesh, (d) reprojected model joints and 
        (e) silhouette reprojection error.
    }
    \label{fig:qualresults_sup}
\end{figure*}

\clearpage
\bibliographystyle{splncs04}
\bibliography{egbib}

{
\setcounter{figure}{0}
\renewcommand\thefigure{\Roman{figure}}
\newpage\cleardoublepage\appendix% updated April 2002 by Antje Endemann
% Based on CVPR 07 and LNCS, with modifications by DAF, AZ and elle, 2008 and AA, 2010, and CC, 2011; TT, 2014; AAS, 2016; AAS, 2020

% \documentclass[runningheads]{llncs}
% \usepackage{graphicx}
% \usepackage{comment}
% \usepackage{amsmath,amssymb} % define this before the line numbering.
% \usepackage{color}
% \usepackage{cleveref}
% \usepackage{booktabs}
% \usepackage{cuted}
% \usepackage{array}
% \usepackage{ragged2e}
% \usepackage{blindtext}
% \setlength{\columnseprule}{0.4pt}
% \usepackage{tabularx}

% INITIAL SUBMISSION - The following two lines are NOT commented
% CAMERA READY - Comment OUT the following two lines
% \usepackage{ruler}
% \usepackage[width=122mm,left=12mm,paperwidth=146mm,height=193mm,top=12mm,paperheight=217mm]{geometry}

% Generic math commands
\def\E#1{E_\textrm{#1}}
\def\W#1{\lambda_\textrm{#1}}
\def\L#1{L_{\textrm{#1}}}

% \begin{document}
% \linenumbers
% \pagestyle{headings}
% \mainmatter
% \def\ECCVSubNumber{1303}  % Insert your submission number here
% \newcommand{\anote}[1]{{\color{red}[#1]}}

% \setlength{\belowcaptionskip}{-10pt}
% % \setlength{\abovecaptionskip}{20pt}

% % I think the pun is good, because let -> left....
\title{Who Left the Dogs Out? \\Supplementary Material} % Replace with your title

\titlerunning{Who left the dogs out?} 
\author{}
\institute{}
% \author{Benjamin Biggs\inst{1} \and
% Oliver Boyne\inst{1} \and
% James Charles\inst{1} \and\\
% Andrew Fitzgibbon\inst{2} \and
% Roberto Cipolla\inst{1}}
% \authorrunning{B. Biggs et al.}

% \institute{
%     Department of Engineering,
%     University of Cambridge,
%     Cambridge, 
%     UK
%     \email{\{bjb56,ob312,jjc75,rc10001\}@cam.ac.uk} \and
%     Microsoft, 
%     Cambridge,
%     UK
%     \email{awf@microsoft.com}}

% % INITIAL SUBMISSION 
% \begin{comment}
% \titlerunning{Who left the dogs out?} 
% \author{Benjamin Biggs\inst{1} \and
% Oliver Boyne\inst{1} \and
% James Charles\inst{1} \and\\
% Andrew Fitzgibbon\inst{2} \and
% Roberto Cipolla\inst{1}}
% %
% \authorrunning{B. Biggs et al.}
% \institute{
%     Department of Engineering,
%     University of Cambridge,
%     Cambridge, 
%     UK
%     \email{\{bjb56,ob312,jjc75,rc10001\}@cam.ac.uk} \and
%     Microsoft, 
%     Cambridge,
%     UK
%     \email{awf@microsoft.com}}
% \end{comment}
%******************

\def\ss#1{\vspace{-0ex}\subsubsection{#1}}

\def\R#1{{\mathbb{R}^{#1}}}
\def\RR#1#2{{\mathbb{R}^{#1 \times #2}}}
\def\posn{\phi}
\def\pose{\theta}
\def\npose{P}
\def\shape{\beta}
\def\scale{\kappa}
\def\trans{t}
\def\betacov{{\Sigma_{\beta}}}
\def\posecov{{\Sigma_{\pose}}}
\def\posemean{{\mu_{\pose}}}
\def\betamean{{\mu_{\beta}}}
\def\nimages{N}
\def\nshape{B}
\def\verts{\nu}
\def\nverts{V}
\def\jointselect{\mathtt{K}}
\def\njoints{J}
\def\f{f}

\begin{comment}
\titlerunning{Who left the dogs out?}
\author{Benjamin Biggs\inst{1}\orcidID{0000-1111-2222-3333} \and
Ollie Boyne\inst{1}\orcidID{1111-2222-3333-4444} \and
James Charles\inst{1}\orcidID{1111-2222-3333-4444} \and
Andrew Fitzgibbon\inst{2}\orcidID{2222--3333-4444-5555}
\and
Roberto Cipolla\inst{1}\orcidID{2222--3333-4444-5555}
%
\authorrunning{F. Author et al.}
\institute{University of Cambridge \and
Cambridge, Tiergartenstr. 17, 69121 Heidelberg, Germany
\email{lncs@springer.com}\\
\url{http://www.springer.com/gp/computer-science/lncs} \and
Microsoft Research, Rupert-Karls-University Heidelberg, Heidelberg, Germany\\
\email{\{abc,lncs\}@uni-heidelberg.de}}
\end{comment}
%******************
\maketitle
% https://www.overleaf.com/project/5e4fcc782813ac000121f3e4
~\vspace{-12mm}\\

\section{Dataset curation}

In this section, we describe our process for  obtaining keypoint and segmentation annotations for the Stanford Dog Dataset~\cite{StanfordDogs}. We submit the entire set of 20,580 dog images to the Amazon Mechnical Turk crowdsourcing platform to obtain a set of 20 keypoint and segmentation masks. We overlay 1 bounding box, provided with the original dataset, on the submitted images to identify the specific dog for the annotators to label. Each image was sent to 3 independent annotators for collecting keypoints and segmentation masks.

\vspace{-1em}
\subsubsection{Keypoints.} 
To identify keypoints, workers were given a list of 20 keypoints to click: 2 per tail, 3 per leg, 2 per ear, nose and jaw. They were additionally asked to provide a visibility flag per point. 

For each keypoint, we process the three clicks to yield a reliable coordinate. From the 3 clicks, we discard clicks that are further than a set tolerance from the mean. If at least 2 clicks remain, we take the mean coordinate as the accepted keypoint position. Otherwise, the point has not been reliably identified between workers, so we set the keypoint as invisible. As described in the main paper, we remove images from train and test splits which have fewer than 8 visible keypoints.

% We rely on the following scheme to determine the optimal position of each keypoint, from the three annotations:

% \begin{enumerate}
%     \item 
% \end{enumerate}

% \begin{enumerate}
% \itemsep0em 
% \item{If If the keypoint was flagged highlighted by three different workers, and all three points are within a certain tolerance of the mean\footnote{Tolerance set at (Image width + Image height)/80, in pixels.}, then accept the mean as the keypoint position. If not, reject the furthest point from the mean, and go to step 2.}

% \item{If the keypoint was highlighted by two different workers, and both points are within a certain tolerance of the mean, accept the mean as the keypoint position. If not, reject the keypoint.}

% \item{If the keypoint was highlighted by only one worker, reject the keypoint.}

% \end{enumerate}

\vspace{-1em}
\subsubsection{Segmentation.}

For each image, each worker $w \in \{w_{1}, w_{2}, w_{3}\}$ submits a binary segmentation mask $\mathbf{A}^{w} \in \mathbb{R}^{H \times W}$. We request a re-labelling for any submissions which fail simple criteria, such as if the highlighted area is below a threshold number of pixels.

% In order to sanitise the dataset and reduce erroneous entries, certain images were rejected or discarded. `Discarded' refers to a submission being discarded from the final dataset, whilst `rejected' means that the submission was deleted from the MTurk dataset, and a new entry requested from the website.\\

% As an initial pass, images were rejected (for which the bounding boxes were insufficiently large - to count null entries. For this, any entry was rejected if,

% \begin{equation}
% \sum\limits_{j=1}^H \sum\limits_{k=1}^W \mathbf{A}^{i,w}_{jk} < 0.01 WH 
% \end{equation}

For each image, we generate the most likely segmentation by comparing submissions across workers. For any two workers $w, w'$ we compute a correlation coefficient:

\begin{equation}
c_{w,w'} = \frac{\sum_i\sum_j \left[ \mathbf{A}^{w} \odot \mathbf{A}^{w'} \right]_{i,j}}{\max\limits_{p = \{w,w'\}} \sum_{i} \sum_{j} \mathbf{A}^p_{i,j}}
\end{equation}

% \begin{equation}
% c_{w,w'} = \frac{\mathbf{A}_{w} \odot \mathbf{A}_{w'}}{\max\limits_{p = \{w, w'\}}\mathbf{A}_{p}}
% \end{equation}

% We remove annotations for which all correlation coefficients are below 80\%. 
Where $\odot$ denotes the element-wise product of the matrices. We remove a worker's segmentation $A^{w}$ if all correlation coefficients $c_{w,w'}$ are below a set threshold. The final binary mask is computed from the remaining submissions:

\begin{equation}
\hat{A}_{i,j} = \begin{cases}
    1, & \text{if } \sum_w A^{w}_{i,j} > 1 \\
    0,              & \text{otherwise}
\end{cases}
\end{equation}

\begin{figure}
\centering
\includegraphics[width=0.95\textwidth]{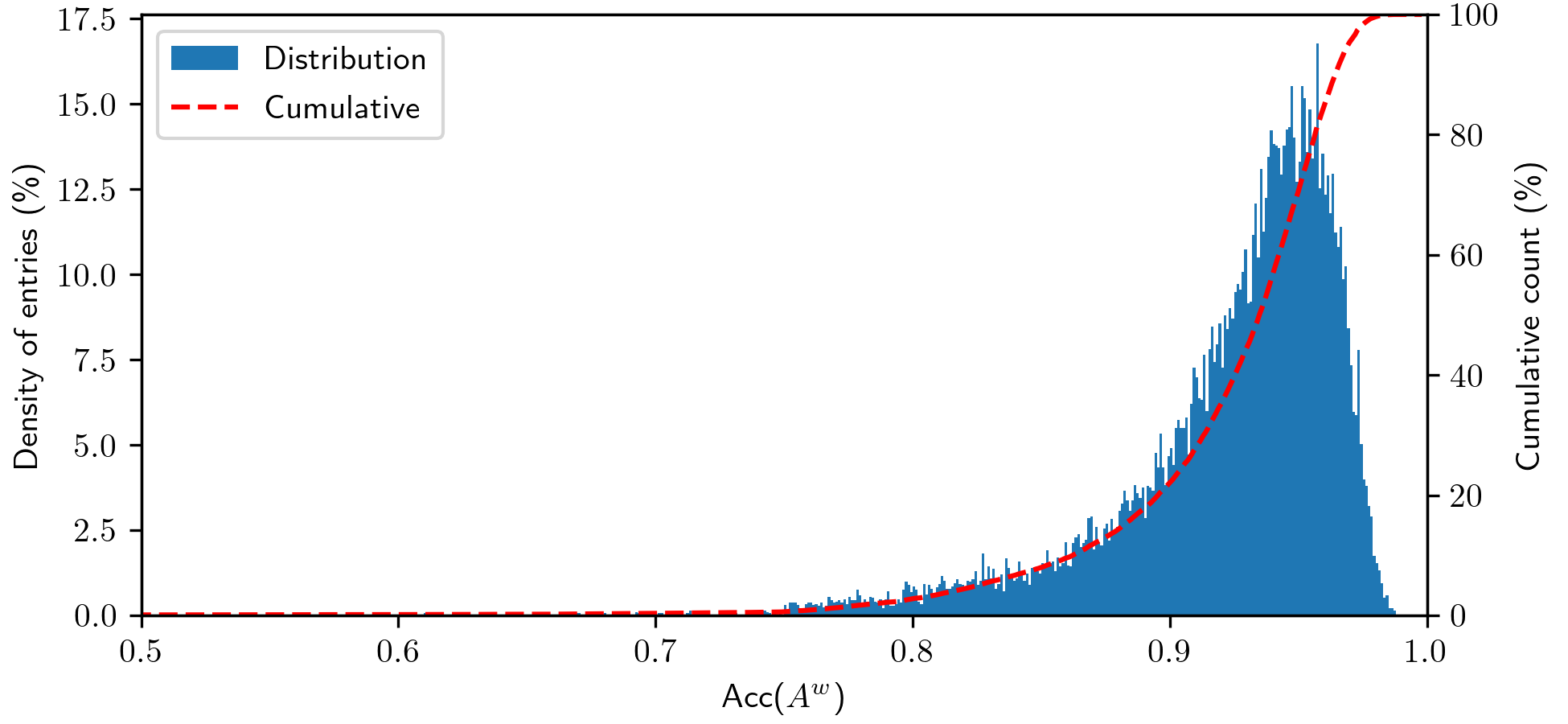}
\caption{Accuracy distribution of all submitted dog segmentations across the entire Stanford Dog Dataset.}
\label{fig:accuracy distribution}
\end{figure}

We can also define the accuracy of a worker's segmentation, as the largest of their correlation coefficients: $\textrm{Acc}(A^{w})=\max_{w'\neq w} \{c_{w,w'}\}$. Figure \ref{fig:accuracy distribution} shows the set of segmentation annotation accuracies over the entire labelled dataset.

% \subsubsection{Filtering the dataset.}

% As described in the paper, we reject images from our train/evaluation splits which based on reliability we deem unsuitable for our task of complete 3D reconstruction. have fewer than 8 identified keypoints find it necessary to reject The Stanford dog dataset contains 20,580 images. Not all of these images are constructive to this dataset - many do not have a lot of the dog in frame, etc. Several stages of refinement of the dataset were made, detailed in Table \ref{tab:filtering}.

% \begin{table}[]
%     \centering
%     \begin{tabular}{c|c}
%          \toprule
%          20,580 &  Stanford dataset\\
%          - (5,...) & Initial filtering based on bbox \\
%          - (5,...) & Images with fewer than 8 identified keypoints \\
%          - (60) & Images with insufficient segmentation accuracy \\
%          - (76) & Images for which more than one keypoint is outside of the segmentation\footnote{(with padding of 5 pixels)}\\
%          \midrule
%          9,647 & Final dataset\\
%          \bottomrule
%     \end{tabular}
%     \caption{Filtering process for final produced dataset}
%     \label{tab:filtering}
% \end{table}

\section{Fitting SMBLD to 3D animation data}

Another method for improving the generalizability of the SMAL model is to improve the 3D shape prior. Such priors are typically used to ensure shape deformation remain within a realistic and anatomically plausible range. Due to the limited diversity of scans used to build the SMAL model, while the shape prior does enforce realism among deformations, it does not allow for a wide enough range to cover the set of dogs in our dataset.

We improve the quality of the prior (and learn a prior over our new scale parameters) by fitting to a set of $13$ artist-designed 3D dog meshes, designed for animation use, which are more varied than the original set. We apply an energy minimization scheme which aligns the SMAL vertices to each scan, under smoothing regularizers.

Recall that SMBLD is adapted from the SMAL~\cite{DBLP:journals/corr/ZuffiKJB16} deformable animal mesh, by including limb scaling parameters. We learn a prior by fitting our SMBLD model, which comprises parameters for pose $\pose$ and shape $\beta$ (the latter of which includes our scaling parameters $\kappa$).

Note that fitting SMBLD to 3D scans is significantly easier than to 2D images, since the complete 3D information of the target mesh is available. In addition, our target meshes are not particularly detailed and are already aligned in T-pose, so we avoid need for a complex alignment technique as discussed in, for example SMPL~\cite{loper2015smpl} or SMAL~\cite{DBLP:journals/corr/ZuffiKJB16}.

We run an energy minimization process to align the SMBLD mesh to the 3D scans, subject to some smoothing regularizers. We minimize the following energy formulation:

\begin{equation}
    \E{opt} = \E{chamfer} + \E{laplacian} + \E{edge} + \E{normal}
\end{equation}
where each of these terms has a scalar weight $\lambda$. We set $\W{chamfer}=\W{edge}=1.0$, $\W{normal}=0.01$ and $\W{laplacian}=0.1$. We run the optimization using SGD, learning rate $1e-4$ for $1000$ iterations.

\vspace{-1em}
\ss{Chamfer energy.} A measure of the average distance between vertices of the SMBLD mesh $V=F_v(\pose, \shape)$, and the target mesh vertices $V'$, when $p$ vertices $v_{i}, v'_{j}$ are sampled from each mesh respectively:
\begin{equation}
        \E{chamfer}(V, V') = \frac{1}{p} \sum_{i=1}^p \min_j^p  \left | v_{i} - v'_{j} \right |
\end{equation}

\vspace{-1em}
\ss{Uniform laplacian energy.} A measure of the mesh smoothness.

\vspace{-1em}
\ss{Edge energy.} This energy is equal to the average edge length across the mesh, and is used to encourage uniform distribution of vertices.

\vspace{-1em}
\ss{Normal energy.} This energy promotes consistency between adjacent faces. It is a measure of the average normal consistency between adjacent faces. For two faces with normals $\mathbf{n_0}$ and $\mathbf{n_1}$, the normal consistency is $1 - \frac{\mathbf{n_0} \cdot \mathbf{n_1}}{\left|\mathbf{n_0}\right|\left|\mathbf{n_1}\right|}$.

At the end of this process, we have a collection of fits $\left|(\pose,\shape)\right|_{\{i=1,...13\}}$ from which we can learn our unimodal pose and shape priors. As discussed, we evenutally use this unimodal shape prior to initialize our mixture shape prior, which is tuned with the expectation-maximization step in the training loop.

% \section{Learning mixture shape prior.}
% This section contains additional detail for how we learn our mixture shape prior, using expectation maximization in-the-loop.

\section{Training procedure}

Recall that the training objective for our end-to-end system for predicting SMBLD parameters consistent with a monocular dog input image is given by:

\begin{equation}
    \L{opt}=\L{joints}+\L{sil}+\L{pose}+\L{shape}+\L{mixture}
\end{equation}

As described in the paper, each loss term is weighted with a scalar $\W{}$ and we train our method in two stages:

\vspace{-1em}
\ss{Stage 1.} We set $\W{joints}=10.0,\W{pose}=1.0,\W{shape}=1.0,\W{sil}=0.0,\W{mixture}=0.0$. We train this stage for 250 epochs, using the Adam optimizer, with learning rate set to $10^{-4}$. 
\ss{Stage 2.} In this stage, we introduce the silhouette loss to encourage a shape alignment between the projected model silhouette and the ground truth annotation. We set $\W{joints}=10.0,\W{pose}=0.5,\W{shape}=0.0,\W{sil}=100.0,\W{mixture}=0.1$. We train this stage for 150 epochs and run the described EM update step every $K=15$ epochs. We selected to use $M=10$ clusters based on a grid search over $M=1,5,10,25$ and comparing IoU. We again use the Adam optimizer, and set the learning rate to $10^{-5}$.

\section{Probability of Keypoints Max (PCK-MAX)} \label{sec:pckmax}

In this section, we compare reprojected 2D joint accuracy using the \emph{PCK-MAX} evaluation metric. This protocol is similar to the Percentage of Correct Keypoints (PCK) metric~\cite{yang2013articulated} used in the main paper by incoporating `invisible' ground-truth points. The standard PCK metric ignores these points, meaning even correct 3D reconstructions will receive no credit. PCK-MAX instead assumes reconstructed 3D points for missing ground-truth data are correct, providing an interesting upper bound. Results are shown in Table~\ref{tab:baselines} and Table~\ref{tab:animalpose}.

\begin{table}[]
{
    \small
    \centering
    \begin{tabular}{@{}lcccccccc@{}}
    \toprule
    \multicolumn{1}{l}{Method} & 
    \multicolumn{1}{c}{Kps} & 
    \multicolumn{1}{c}{Seg} & 
    \multicolumn{5}{c}{PCK-MAX @ 0.1} \\
    \multicolumn{3}{c}{} &
    \multicolumn{1}{c}{Avg} &
    \multicolumn{1}{c}{Legs} &
    \multicolumn{1}{c}{Tail} &
    \multicolumn{1}{c}{Ears} &
    \multicolumn{1}{c}{Face} \\
    \midrule
    \rowcolor{comp} 3D-M~\cite{DBLP:journals/corr/ZuffiKJB16} & Pred & Pred & 67.1 & 65.7 & 79.5 & 54.9 & 87.4  \\
    \rowcolor{notcomp} 3D-M & GT & GT & 72.6 & 69.9 & 92.0 & 58.6 & 96.9 \\
    \rowcolor{notcomp} 3D-M & GT & Pred & 72.6 & 70.2 & 91.5 & 58.1 & 96.9 \\ 
    \rowcolor{notcomp} 3D-M & Pred & GT & 67.4 & 66.0 & 79.9 & 55.0 & 88.2 \\ 
    \hline
    \rowcolor{comp} CGAS~\cite{biggs2018creatures} & CGAS & Pred & 43.7 & 46.5 & 64.1 & 36.5 & 21.4  \\
    \rowcolor{notcomp} CGAS & CGAS & GT & 43.6 & 46.3 & 64.2 & 36.3 & 21.6 \\
    \hline
    \rowcolor{comp} 3D-M + scaling & Pred & Pred & 69.6 & 69.4 & 79.3 & 56.5 & 87.6 \\
    \rowcolor{comp} 3D-M + scaling + EM prior & Pred & Pred & 71.6 & 71.5 & \textbf{80.7} & 59.3 & 88.0 \\
    \hline
    \rowcolor{comp} \textbf{Ours} & --- & --- & \textbf{75.7} & \textbf{75.0} & 77.6 & \textbf{69.9} & \textbf{90.0} \\
    \bottomrule 
    \end{tabular}
    \vspace{1em}
    \caption{\label{tab:baselines}\textbf{PCK-MAX baselines.} PCK-MAX scores are shown for SOTA methods under varying conditions. Directly comparable baseline methods (requiring only an input image) are highlighted. \emph{Pred} keypoints generated with Hourglass-Net~\cite{newell2016stacked} and segmentations with DeepLab v3+~\cite{journals/corr/ChenPK0Y16}. 3D-M/CGAS are also analysed when they have access to ground-truth keypoints and/or segmentation masks. We also analyse adding this paper's innovations (scale parameters and EM prior) to the 3D-M method~\cite{DBLP:journals/corr/ZuffiKJB16}.}
}
\end{table}

\begin{table}[!htbp]
    \parbox{.45\linewidth}{
        % \3D-Ml
        \centering
        \begin{tabular}{@{}lccccc@{}}
        \toprule
        \multicolumn{1}{l}{Method} & 
        \multicolumn{5}{c}{PCK-MAX @ 0.1} \\
        \multicolumn{1}{c}{} &
        \multicolumn{1}{c}{Avg} &
        \multicolumn{1}{c}{Legs} &
        \multicolumn{1}{c}{Tail} &
        \multicolumn{1}{c}{Ears} &
        \multicolumn{1}{c}{Face} \\
        \midrule
        3D-M~\cite{DBLP:journals/corr/ZuffiKJB16} & 69.1 & 60.9 & 83.5 & 75.0 & 93.0 \\
        % \hline
        \textbf{Ours} & \textbf{73.8} & \textbf{65.1} & \textbf{85.6} & \textbf{84.0} & \textbf{93.6} \\
        \bottomrule
        \multicolumn{6}{c}{} \\
        \multicolumn{6}{c}{}
        % \textbf{Ours} & \textbf{66.9} & \textbf{73.8} & \textbf{65.1} & \textbf{85.6} & \textbf{84.0} & \textbf{93.6} \\
        % \textbf{Ours} & \textbf{66.9} & \textbf{73.8} & \textbf{65.1} & \textbf{85.6} & \textbf{84.0} & \textbf{93.6}
        \end{tabular}
        \vspace{1em}
        \caption{
            \label{tab:animalpose}
            \textbf{PCK-MAX Animal Pose dataset~\cite{animalpose}}. Evaluation on recent Animal Pose dataset with no fine-tuning to our method nor joint/silhouette predictors used for 3D-M.}
    }
    \hfill
    \parbox{.45\linewidth}{
        % \small
        \centering
        \begin{tabular}{@{}lccccc@{}}
        \toprule
        \multicolumn{1}{l}{Method} & 
        \multicolumn{5}{c}{PCK-MAX @ 0.1} \\
        \multicolumn{1}{c}{} &
        \multicolumn{1}{c}{Avg} &
        \multicolumn{1}{c}{Legs} &
        \multicolumn{1}{c}{Tail} &
        \multicolumn{1}{c}{Ears} &
        \multicolumn{1}{c}{Face} \\
        \midrule
        \textbf{Ours} & \textbf{75.7} & \textbf{75.0} & \textbf{77.6} & 69.9 & 90.0 \\
        $-$EM & 74.6 & 72.9 & 75.2 & \textbf{72.5} & 88.3 \\
        $-$MoG & 74.9 & 74.3 & 73.3 & 70.0 & \textbf{90.2} \\ 
        $-$Scale & 72.6 & 72.9 & 75.3 & 62.3 & 89.1 \\
        \bottomrule 
        \end{tabular}
        \vspace{1em}
        \caption{\label{tab:ablation}\textbf{PCK-MAX ablation study.} Evaluation with the following contributions removed: (a) EM updates, (b) Mixture Shape Prior, (c) SMBLD scale parameters.}
    }
\end{table}

}
\end{document}